\RequirePackage{snapshot}
\documentclass[conference,10pt,letterpaper]{IEEEtran}

\usepackage{bbm}
\usepackage{graphicx}
\usepackage{verbatim}
\usepackage{url}
\usepackage{amsmath}
\usepackage{amsfonts}
\usepackage{amssymb}
\usepackage[latin1]{inputenc}
\usepackage{delarray}
\usepackage{subfigure}
\usepackage{setspace}
\usepackage{paralist}
\usepackage{enumerate}
\usepackage{hyperref} 
\usepackage{color}
\usepackage[font=small]{caption} 
\IEEEoverridecommandlockouts

\usepackage{accents}
\newlength{\dhatheight}
\newcommand{\doublehat}[1]{%
    \settoheight{\dhatheight}{\ensuremath{\hat{#1}}}%
    \addtolength{\dhatheight}{-0.35ex}%
    \hat{\vphantom{\rule{1pt}{\dhatheight}}%
    \smash{\hat{#1}}}}

\graphicspath{{./Figures/}}

\newtheorem{example}{Example}[section]

\newcommand{\R}{\mathbb{R}}
\newcommand{\ind}{\mathbf{1}}
\newcommand{\md}{\mathrm{d}}
\usepackage{accents}

\newcommand{\E}{\mathbf E}

\newcommand{\remove}[1]{}
\newcommand{\bfS}{\mathbf{S}}
\newcommand{\bfSS}{\mathbf{S\hspace{-0.08em}2}}
\newcommand{\mytop}{{\!\top}}

\newcommand{\exclude}[1]{}
\allowdisplaybreaks

\begin{document}
\title{Statistical learning of geometric characteristics of wireless networks
\thanks{{\bf A. Brochard} is with the {\em Huawei R\&D} France and {\em Inria/ENS}, Paris, France;  {\bf B. B{\l}aszczyszyn} is with {\em Inria/ENS}, Paris, France; {\bf S. Mallat} is with {\em College de France} and {\em ENS} Paris, France;
{\bf S. Zhang} is with {\em Peking University}, Beijing.}\vspace{-1ex}
}
\author{Antoine Brochard, Bart{\l}omiej B{\l}aszczyszyn, St{\'e}phane  Mallat, Sixin Zhang}

\maketitle

\begin{abstract}
Motivated by the prediction of cell loads in cellular
networks, we formulate the following new, fundamental problem of {\em
 statistical learning of geometric marks of  point processes}: An unknown
{\em marking function}, depending on the geometry of  point patterns,
produces characteristics (marks) of the points.  One aims at learning
this function from the examples of marked point patterns
in order  to predict the  marks of new point patterns.
To approximate (interpolate)  the marking function,
in our baseline approach, we build a statistical  {\em regression} model
of the marks  with respect some {\em local} point distance representation.
In  a more advanced approach, 
we use a {\em global} data representation
via the {\em scattering moments} of random measures, 
which  build  informative and stable to
deformations  data representation,  already  proven useful
in image analysis and related application domains.
In this case, the regression of the scattering moments of the
marked point patterns with respect to the non-marked ones is  combined
with the numerical solution of the {\em inverse problem}, where the
marks are recovered from the estimated scattering moments.
Considering  some simple, generic marks, often appearing in the
modeling of wireless networks, such as the shot-noise values, nearest
neighbour distance, and some characteristics of the Voronoi cells,
we show that the scattering moments can capture similar geometry
information as the baseline approach,
and can reach even better
performance, especially for non-local
marking functions.  Our results motivate  further
development of statistical learning tools for  stochastic geometry
and  analysis of wireless networks, in particular to predict
cell loads in cellular networks from the locations of base
stations and traffic demand.
\end{abstract}

\section{Introduction}
\label{s.introduction}

Design,  performance evaluation and control of wireless networks 
are facing rapid increase in complexity, due to more and more dense deployment of
the classical cellular networks
and the advent  of the Internet of things.
It is clear that  engineering of these networks  needs to make
more systematic use of  the data  massively collected in operational
conditions, thus opening this domain to possible applications of
machine learning methods.

While advanced signal processing techniques at the link layer already integrate elements
of artificial intelligence  (like belief propagation in Bayesian
networks  for low-density parity-check and turbo codes)
it is more seldom to see them used at higher levels, in
particular at the network layer. The data corresponding to this layer
(e.g.  base station locations, their characteristics and
performance metrics,  user distribution and  QoS metrics)
have geometric structure, reflecting (usually
two-dimensional) geographic network deployment.
It is thus natural to ask questions regarding 
pertinence of the rapidly developing machine learning tools for 
image analysis and related fields to  this new domain of
applications.

\subsection{Statistical learning of marks of  point processes}
\label{ss.Intro-problem}
In this paper  we pose and address  to the following
fundamental problem of  statistical learning of marks of  point processes:
 {\em An unknown marking function depending on the relative locations of
points  produces characteristics of these points, called marks.
One aims at learning this function from the examples of point patterns
with observed marks  in order to predict unknown marks
for new point patterns.}

\begin{figure}[t]
  \begin{center}
\vspace{-3ex}
    \centerline{\includegraphics[width=0.6\linewidth,height=0.3\linewidth]{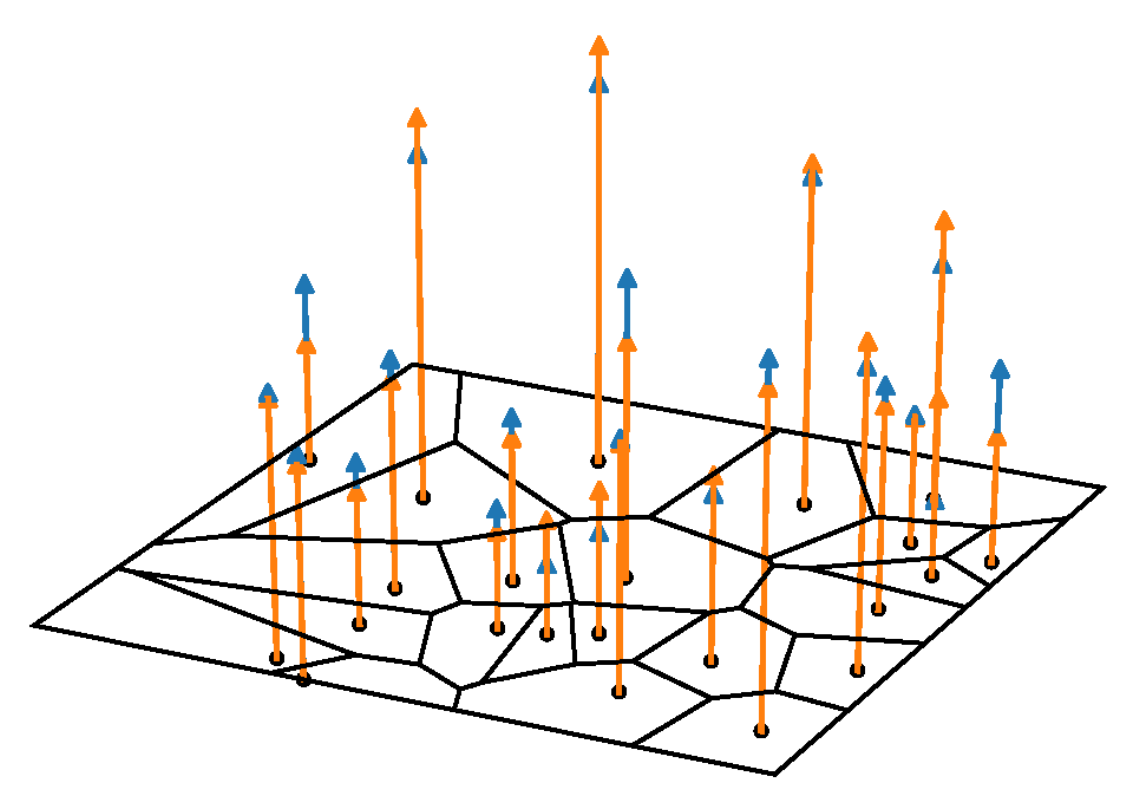}}
  \vspace{-1ex}
 \caption{Exact (blue) and reconstructed (orange) marks being the surface areas
 of the Voronoi cells.}
\label{fig.visualisation}
  \end{center}
  \vspace{-6ex}
\end{figure}

 Our particular motivation comes from the problem of  learning of the
  dependence of the cell loads in wireless networks 
 on  the geometry of base stations
 (and possibly of the traffic demand) directly from real data
 collected in the existing, operational networks,
 to predict loads of base stations for different
 base station positioning  and/or different traffic demand;
 see more details in Section~\ref{ss.Related-works}.
In this founding work, we consider some simple,
generic marks, often appearing in this context. They are produced by the standard
shot-noise interference model,  the nearest neighbour
distance, and some characteristics of the Voronoi cells (cf
Figure~\ref{fig.visualisation}).
Our goal is to understand  the  amenability of marks
representing different types of dependence on point patterns
to the proposed statistical learning approach.

\subsection{Learning via local or global geometry representation}
The learning (interpolation) of the marking function
in the original problem 
does not seem amenable to any direct statistical approach,
due to the structure  of the space of marked point measures.
To overcome this difficulty, we need some suitable 
representation of the geometry of point patterns,
and  we propose two approaches in this regard.

As a baseline, we propse to  estimate the mark of each point using
the statistical regression model based on the {\em local distance
  matrix} of a  suitable chosen vicinity of this point.
In this relatively simple approach,  the training data set consists
of an ensemble of central, marked points   surrounded by
some number, say $K$,  of their (non-marked) neighbours. If no a'priori information regarding the
dependence region of the marking function is available, one needs to
choose~$K$  using cross validation.  Observe that  
the dimension of the domain of the interpolated marking function
(local distance  matrix) increases as $K^2$.
For highly non-local marking functions (e.g.  related to power-law
shot-noise function),  one   might need to take $K$ large,
proportional to the total number of points, making this baseline
approach not efficient.

When the marks have non-local dependence, 
or when we have no prior knowledge of the dependency range,
we further propose
to use the {\em scattering moment} representation~\cite{gscatt} to capture 
the geometry of marked point patterns.  It is a discrete family of  nonlinear and
noncommuting operators, computing at different scales the  modulus of
a wavelet transform of the one- or higher-dimensional
signal (e.g. image).  Applied to the signal observed in a finite window, they are  proven to be
Lipschitz-continuous with respect smooth (class $C^2$) signal diffeomorphisms. As the
window size increases, they  become invariant with respect to the
translation of the signal. They can be also made rotation invariant.
These invariance and stability properties
make them useful in signal processing, in particular in relation to statistical learning. Indeed, if the information content of  an
image typically is not (strongly) affected by translations,
rotations, and  small deformations, similar properties of the signal
representation allows one to capture this content in a more concise
way, and hopefully learn its intrinsic structure  from a smaller number
of signal samples. The pertinence of this approach has already been demonstrated  in
various contexts.

\begin{figure}[t]
  \vspace{-5ex}
  \begin{center}
   \includegraphics[width=0.8\linewidth,height=0.4\linewidth]{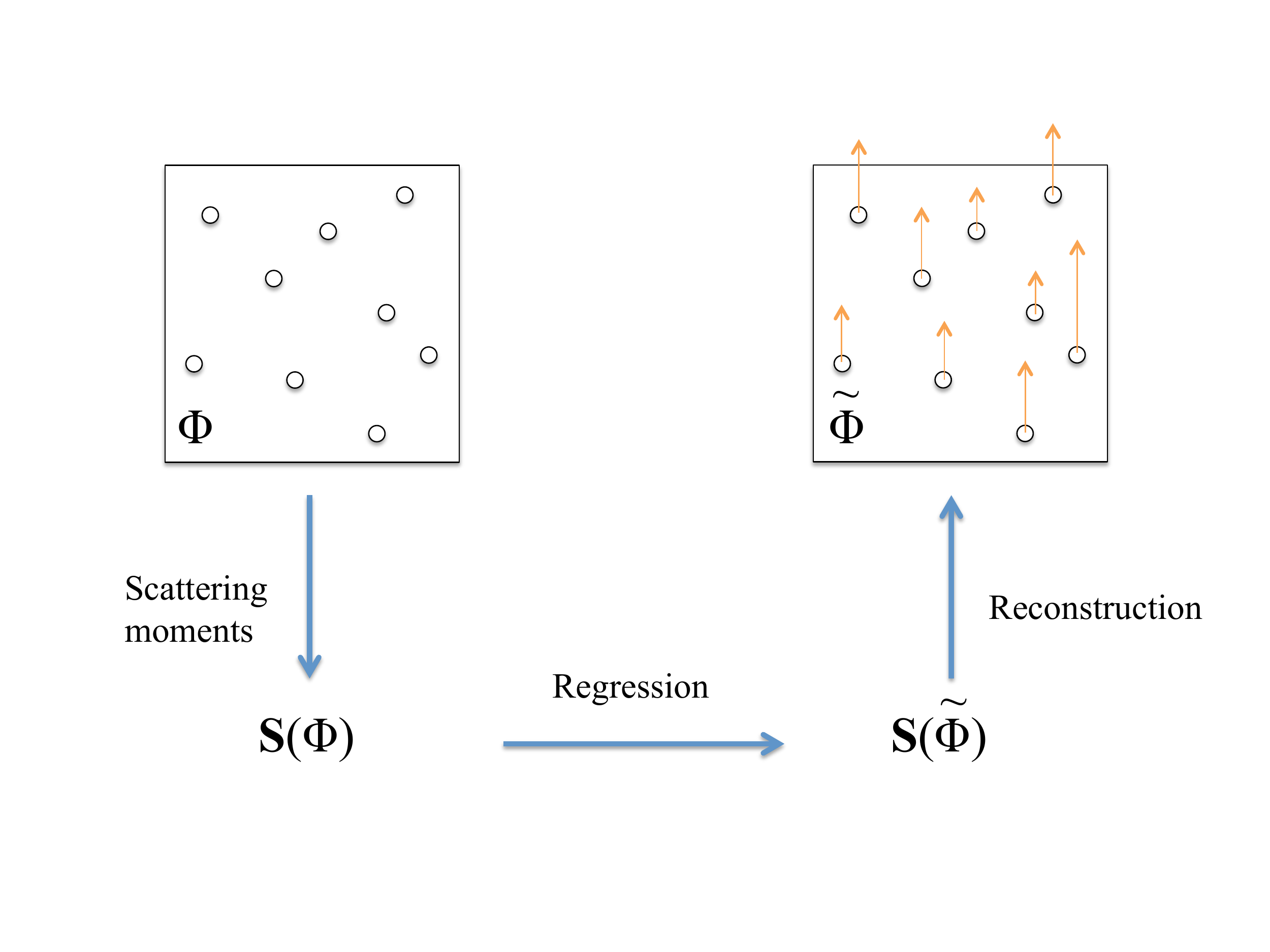}\\[-5ex]
 \caption{Reconstruction of geometric marks via scattering moments}
\label{fig.schema}
  \end{center}
  \vspace{-4ex}
\end{figure}

Using the scattering moments to represent marked point processes,
our  learning
problem can be addressed in the following two steps (also depicted on
Figure~\ref{fig.schema}):
\begin{itemize}
\item We build a statistical regression model on scattering
 moments of the marked point patterns
   with the explanatory variables being the  scattering moments  of the
   non-marked point patterns.
 This model is   computed on the training  data consisting of point
 patterns with observable marks and is meant to be used to  
 estimate
the  (marked) scattering moments of new  point patterns, for which marks are
not observed. 
 \item We estimate (reconstruct)  the marks of new  point patterns, for which marks
  are not observed,  from their  estimated  scattering moments.
  It is a non-convex  optimization
  problem. We solve it numerically using L-BFGS-B algorithm~\cite{lbfgs,lbfgs2}.
 \end{itemize}

This approach scales much better with the total number of points. Indeed, the number of scattering moments capturing the
global geometry grows logarithmically with this number. However, it may suffer from
possible errors introduced in the reconstruction phase
(absent in the benchmark approach).
The overall benefits of the scattering approach become significant for highly non-local marking
functions. 

In general, the quality of the regression of the scattering moments 
depends on the sensitivity  of the marking function to small point pattern deformations ---  a concept not yet well
understood in point process literature (e.g. the nearest neighbour
distance seems to be more sensitive than shot-noise). 
On the other side, in  the reconstruction phase (recovering of  marks
from the true or estimated scattering moments) 
significant  errors  consist in swaping  a large and
a small  mark of two neighbouring points (not  leading to a significant
modification of the considered first scattering moments).
The quality of this phase of the approach depends thus on the existence of clusters of points in  point patterns --- a problem already
studied in the literature; cf Section~\ref{ss.Related-works}. The Poisson point process,
considered in this paper, exhibits  a baseline  type of clustering.

\subsection{Related works}
\label{ss.Related-works}
Stochastic-geometric study of cellular networks
expands rapidly in recent years primarily through
analytic results regarding Poisson network
models~\cite{blaszczyszyn2018stochastic}.  
Performance evaluation of operational wireless cellular networks, in particular
the quality of service perceived by users in  function of the traffic
demand, is a complex problem  involving stochastic and geometric 
modelling of several network layers.  A key element of this problem is
the analysis of the cell loads, which non trivially depend 
on the geometric configuration of serving base stations and their
traffic demands, and   capture in  concise way the quality of service offered by the
individual cells. A detailed physical    model of cell loads 
was  proposed~\cite{siomina2012analysis} and
revisited in~\cite{blaszczyszyn2014user,blaszczyszyn2016spatial,blaszczyszyn:hal-01824986}
including the  validation  with respect to some data collected in operational
networks.

Recently, the  prediction of the  cell loads has been
the  subject of a  machine learning study published  in~\cite{awan2018robust},
though   {\em not} as the  geometric problem posed in our paper.
Indeed, in this earlier  paper the number and the  locations  of base stations
are {\em fixed}, and the problem consist in learning the loads  of this
given configuration of stations in function of their traffic demands.
This (non-geometric) problem is cast and solved in the framework of
monotone interpolation of Lipschitz functions.

The problem formulated and studied in our paper allows one  to address
the completely orthogonal question of learning of the cell loads,
given  constant traffic demand, in function of the geometry of the
network.   Combining the two approaches, left for the future work,
will allow one   to predict the performance of new
geometric configurations on base stations and traffic demand by collecting 
the cell load data in existing,  operational  networks,
thus offering an alternative to building and solving complex physical
models.

Observe that, even  for a fixed number of base stations,
the  two dimensional  
location of each base station on the plane makes   our
 geometric argument of the marking function have two times higher dimension than the
 one considered in~\cite{awan2018robust}.
 Using  wavelet scattering transforms  allows one  to leverage geometric invariants, such as
translation and rotation invariance with respect to the locations of
stations, and thus significantly reduce the representation dimension.  Scattering moments have been  proven useful in this context
in a number of tasks such
as quantum molecular energy prediction~\cite{multi} or texture classification~\cite{srqe,text}.

Geometric marks (called also score functions) have received a lot of attention in stochastic geometry and spatial statistics, where they represent some interaction of a given point  with the whole point pattern. If this interaction is local in some sense,
and the underlying point process  exhibits some decay of correlations, then it is possible to establish asymptotic results
(including central limit theorems)
regarding the sums of the geometric marks observed in increasing  window; cf  e.g.~\cite{penrose2003weak,baryshnikov2005gaussian,blaszczyszyn2016limit}. These results can be used  to study large-scale asymptotic of the scattering moments as in~\cite{scatt}.

It was mentioned in the introduction that the quality of  reconstruction of the marks depends on whether point process is regular or clusters its points.
These notions are formalised in~\cite{dcx-clust,blaszczyszyn2015clustering},
and it is argued that Poisson point process exhibits some reference type of
point clustering, to which more regular point processes (e.g. determinantal ones) and
more clustering ones (e.g. permanental and Cox processes) can be
compared in terms of the performance of different geometric characteristics.

\medskip
\paragraph*{The remaining part of this paper is organized as follows}
In Section~\ref{s.Marks-model} we describe several models of geometric marks
and formulate the problem of their statistical learning.
In Section~\ref{s.Learning} we present our main approach proposed to
solve this problem, based on scattering transforms of marked point
processes. We also describe  the  benchmark approach.
The numerical results of both approaches  applied to the considered
mark models 
are presented in Section~\ref{s.Results}.

\section{Geometric characteristic of wireless networks}
\label{s.Marks-model}
In stochastic-geometric modeling of wireless networks, one usually
 represent locations of transmitters/receivers as {\em points} of a point
 process, and  their characteristics as {\em marks}.
 These characteristics can depend not only  on the given
 transmitter/receiver (as e.g. the type of the base station in
 heterogeneous cellular networks or the local density of users in the
 vicinity of this station)  but also on (at least)  local geometry
 of the network 
 (as e.g. the surface of the cell served by the base station, the extra-cell
 interference). 
These marks, called in what follows {\em geometric marks},  are in the center of our interest in this paper.

\subsection{Geometric marks}
\label{ss.GM-model}
   More formally,  let $\Phi=\sum_i\delta_{X_i}$ be a simple
 ($X_i\not=X_j$ for $i\not=j$) , stationary (having distribution
 invariant by any translation) point process on the
Euclidean plane~$\R^2$. Recall, $\Phi$ is a random object on the
space  $\mathbb{M}$ of locally finite counting measures on $\R^2$
with a suitable $\sigma$-field; cf~\cite{daley2007introduction}. 
Consider a (measurable) {\em marking function} $m=m(x,\phi)$ defined on
$\R^2\times\mathbb{M}$. Assume $m$ is invariant with respect to
translations on the plane, i.e., $m(x+a,\phi+a)=m(x,\phi)$ for any
$a\in\R^2$, where $\phi+a$ is the translation of the counting measure
(of its atoms) by the vector~$a$.
We will call $M_i:=m(X_i,\Phi)$   {\em geometric mark} of the point~$X_i$ of~$\Phi$.
Note that $\Phi^m:=\sum_i\delta_{(X_i,M_i)}$ is a stationary
marked point process.

In what follows we briefly remind a few basic examples of geometric
marks  that will be
considered in the remaining part of this paper.
For better understanding of the context  we interpret  points $X_i$ as locations of base stations
of some cellular network.

\subsubsection{Shot-noise}
\label{sss.SN-model}
For $x\in\R^2$ and  $\phi=\sum_{i}\delta_{x_i}\in\mathbb{M}$, let
$m(x,\phi):=\sum_{i}\ind(x\not=x_i)\ell(|x-x_i|)$ be a {\em shot-noise}
functional with some non-negative response function $\ell$ of the
distance $|x-x_i|$ between $x$ and $x_i\in\phi$.
Taking $\ell(r)=r^\beta$ 
to be the standard power-law path-loss model with the path-loss exponent
$\beta>2$, we recognize in   $S_i:=m(X_i,\Phi)$  the total power
(usually interpreted as {\em interference})
received at the station $X_i$ from other stations of the network, all
transmitting with a unit power.

\subsubsection{Nearest neighbour distance}
Let $R_i:=\min\{|X_i-X_j|:x_j\in\Phi,X_i\not=X_i\}$
be the the distance from $X_i$ to its nearest
point in $\Phi$. Similarly one can define distances to successive
neighbours of $X_i$ in $\Phi$.

\subsubsection{Voronoi cell surface area and  moment of inertia}
Denote by  $V_i:=\{y\in \R^2: |y-X_i|\le \min_{X_j\in\Phi}|X_j-X_i|\}$
the Voronoi cell of the point $X_i$ in $\Phi$. It is the polygon
representing all locations  on the plane closer to $X_i$ than to any
other point in $\Phi$. This is a fundamental  cell
model for cellular networks.
We will consider two basic numerical characteristics of the Voronoi
cells: the surface area  $A_i:=\int_{V_i}1\,\md y$ and the moment of
inertia $I_i:=\int_{V_i}|y-X_i|^2\,\md y$.
The marks $A_i$ and $I_i$  can be interpreted as very simple proxies to
the traffic demand and the cell load  of the station $X_i$
(the latter assuming user peak-bit rate satisfies the inverse square law with
respect to its serving station;
cf~\cite[Section~4.1.8]{blaszczyszyn2018stochastic}). 

\subsubsection{Voronoi shot-noise}
By this we call the marks  $Z_i$ defined as follows:
$Z_i:=\sum_{j}\ind(X_j\not=X_i)\ell(|X_j-X_i|)A_j$.
Having interpreted shot-noise $S_i$ of Section~\ref{sss.SN-model}
as the interference received at the station  $X_i$ from all other
stations transmitting a unit power, $Z_i$ can be also interpreted
as the  interference, however with the  stations transmitting
the signals with the power proportional to their Voronoi cells.
(Similar kind of dependence, with more complicated expressions,
can be recognized the cell load model of~\cite{siomina2012analysis}.)

\subsection{Problem formulation}
\label{ss.Problem}
Let us now formulate the main  problem studied in this paper, that is
the problem of {\em learning of geometric marks}.
Suppose the marking function $m$ is not known explicitly.
One observes only some 
realizations of the marked point process $\Phi^m$
with points restricted to some finite observation window $W$. Denote these
realizations by $\phi_k^m=\sum_{i}\delta_{(x_i(k),m_i(k))}$, with
$x_i(k)\in W$, $k=1,\ldots$. They form a {\em training set} of data.
The problem consists in  learning  the function
$m$ from the training set so as to be able  to calculate approximations of the unobserved  marks $m_i=m(x_i,\phi)$
for a new realization $\phi=\sum_i\delta_{x_i}$  of (only points) of the point process $\Phi$.

\section{Statistical learning of geometric marks}
\label{s.Learning}
In this section we propose a solution to 
the learning  problem 
formulated in Section~\ref{ss.Problem}. Our  main  approach, described
in Section~\ref{ss.Learnig-via-scattering},
uses a wavelet based  representation of marked point patterns. It is 
presented in  Section~\ref{ss.Scattering}.
Finally, in Section~\ref{ss.Learnig-benchmark}
we  describe an alternative method, based on a distance matrix
representation of  point neighbourhoods,  used as a benchmark approach
for our numerical study in Section~\ref{s.Results}.

\subsection{Scattering moments of marked point processes}
\label{ss.Scattering}
Following~\cite{scatt}, now  we  shall  briefly present the 
wavelet scattering transform of marked  point processes on the plane~$\R^2$.

Let $\psi$ be a continuous, bounded, complex function on $\R^2$ of
zero average $\int_{\R^2}\psi(x)\,\md x=0$ 
and such that $|\psi(x)|=O(|x|^{-2})$ for $|x|\to\infty$.
It is customary to  normalize it  so that
$\int_{\R^2}|\psi(x)|\,\md x=1$.
We call $\psi$ (two-dimensional) {\em mother  wavelet}.

\begin{example}
  \label{example:Morlet}
  Two dimensional {\em Morlet wavelet} has
the following form 
$\psi(x)=\exp(i\; \omega\!\cdot\! x)\exp(-|x|^2/2)$, where $i$ is the
imaginary unit and  $\omega\!\cdot\! x$ is the scalar product of 
some nonzero vector parameter $\omega\in\R^2$, called spatial
frequency, with $x\in\R^2$. Note that $\psi$ it is not normalized
and, moreover, it  is  only approximately  zero average  when $|\omega|$ is large
enough;  typically  $|\omega|\ge 5.5$; cf~\cite{antoine1993image}.

\end{example}

Consider a family of  {\em dilated}, and {\em rotated wavelets} 
$\psi_{(j,\theta)}$: for $j \in
\mathbb{Z}=\{\,\ldots,-1,0,1,\ldots\,\}$ and $\theta \in [ 0, 2\pi)$
  let $\psi_{(j,\theta)}(x):=2^{-2j}\psi(2^{-j}r_{-\theta}x)$, where
  $r_{\theta}x$ deontes the rotation of $x\in\R^2$ by the angle
  $\theta$ with respect to the origin on the plane (and the factor $2$ in
  the first exponent corresponds to the dimension~$2$).

Let $\tilde\Phi=\sum_{i}\delta_{(X_i,U_i)}$ be a simple stationary marked
point process with points $X_i\in\R^2$ and some real  marks $U_i\in\R$.
The {\em wavelet transform} of $\tilde\Phi$ at scale $2^j$ and angle
$\theta$, $j\in\mathbb{Z}$, $\theta\in[0,2\pi)$,
 is defined as the following (stationary, complex) random filed on $\R^2$:
  \begin{align}
  \tilde\Phi \star \psi_{(j,\theta)}(x) &:= \int_{\R^2\times
    \R}\psi_{(j,\theta)}(x-y)\,u\tilde\Phi(\md(y,u)) \label{eq:PPwavelet-intergal}\\
    &= \sum_{i}U_i \psi_{(j,\theta)}(x-X_i)\,.\label{eq:PPwavelet}
    \end{align}
The  wavelet transform $\Phi \star \psi_{(j,\theta)}$ of a
(non-marked) point process $\Phi=\sum_{i}\delta_{X_i}$ is defined
by~\eqref{eq:PPwavelet} with unit marks $U_i\equiv 1$.

The {\em first order scattering moments} $\bar S\tilde
\Phi({j,\theta})$ of $\tilde\Phi$,  at scale $2^{j}$, $j\in\mathbb{Z}$
and angle $\theta\in[0,2\pi)$, are defined
  as the expectation of the modulus of the corresponding wavelet
  transform (without loss of generality evaluated at the origin)
\begin{equation}\label{eq:S1}
  \bar S\tilde\Phi({j,\theta}):=\E[|\tilde\Phi \star \psi_{(j,
      \theta)}(0)|]\,,
\end{equation}
and similarly for the non-marked point process $\Phi$.
 
The {\em second order scattering moments}
$\bar S\tilde \Phi({j_1,\theta_1,j_2,\theta_2})$, $j_i,j_2\in\mathbb{Z}$,
$\theta_1,\theta_2\in[0,2\pi)$,
  are  defined by induction as  the first order scattering moments 
 at scale $2^{j_2}$ and angle $\theta_2$
  of the random measure  on $\R^2$ admitting the modulus  $|\tilde\Phi \star \psi_{(j_,\theta_1)}(x)|$
  of the wavelet transform at scale $2^{j_1}$ and angle
$\theta_1$ for density  (this measure replaces the projecton of the
  point measure  
$u\tilde\Phi(\md(y,u))$ on $\R^2$ in~\eqref{eq:PPwavelet-intergal}) 
\begin{equation}\label{eq:S2}
    \bar S\Phi({j_1,\theta_1}, {j_2,\theta_2}):=
  \E[||\Phi \star \psi_{(j_1,\theta_1)}|\star
    \psi_{(j_2,\theta_2)}(0))|]\,.
  \end{equation}
Higher order scattering moments are defined by induction
(but they are not used in this paper).
Scattering moments have not yet been fully theoretically studied
 on the ground of the theory of point processes.
Some asymptotic properties at  small and large scale
($j\to -\infty$ and $j\to\infty$ respectively) can be established,
extending results presented in~\cite{scatt}.
\footnote{
  In particular, it is not known to what extend the scattering moments
characterize the distribution of the  point process.
Note that higher order factorial  moment measures
(their densities, if exist, are called the correlation
functions)  are known to characterize the distribution of the simple
point process having finite exponential moments.
The following two observations can be made regarding the relation
between the  correlation functions and  the scattering moments.

Note that the wavelet transforms~\eqref{eq:PPwavelet} are
linear functions (shot-noise functionals) of the (marked) point
process; in other words they are  first-order
$U$-statistics of $\tilde\Phi$; cf~\cite[Section~12.3]{last2017lectures}.
The fact that scattering moments  $\bar
S\Phi(j,\theta)$ are defined via the moduli   of the wavelet transforms
makes them dependent on all higher-order correlation
functions of $\Phi$: the factorial moment expansions~\cite{blaszczyszyn1995factorial,blaszczyszyn1997note}
of   $\bar
S\Phi(j,\theta)$ involve all moment measures
(in contrast to  the square norm of the wavelet transforms,
which can be represented using the first and second-order
$U$-statistics and thus their  expansion involves only  the first two
correlation functions).
Consequently,  already the first scattering moments
are supposed to capture more information regarding the  intrinsic
dependence of the points
than just the  pairwise correlations. We shall see in our numerical
study that this information allows one for an  efficient
recovery of some geometric marks which do depend on higher order
correlations, e.g, the Voronoi cell  characteristics.

Even if we do not know whether the (higher-order) scattering moments
characterize the distribution of the point processes,
being a discrete family of {\em numbers} (provided a suitable
discretization of $\theta\in[0,2\pi)$)  
they are supposed to capture  the point correlations   in a more concise  way than the
higher-order correlation {\em functions}. In practice,  a reasonable estimation
of the scattering moments requires much less examples of point
patterns than the estimation of the moment mesures.}

\subsubsection{Empirical scattering moments}
\label{sss.Scat-Estimate}
For a given realization $\tilde\phi$ of $\tilde\Phi$ observed in a finite window~$W$,
{\em empirical  scattering moments} $\hat S\tilde\phi({j,\theta})$  and
$\hat S\tilde\phi({j_1,\theta_1}, {j_2,\theta_2})$
are obtained replacing the expectation $\E[\cdots]$
in~\eqref{eq:S1} and~\eqref{eq:S2} by the empirical averaging of the fileds
$|\tilde\phi \star \psi_{(j,    \theta)}(x)|$ and 
$||\phi \star \psi_{(j_1,\theta_1)}|\star
    \psi_{(j_2,\theta_2)}(x))|$,  resepectively, over $x\in W$.
     When~$W$ increses suitable to the whole plane,
  these (empirical) moments become  asymptotically non-biased estimators
  of  $\bar S\Phi({j,\theta})$ and $\bar
  S\Phi({j_1,\theta_1},{j_2,\theta_2})$, respectively, 
  provided
  $\tilde\Phi$ is ergodic, at least when the mother wavelet has finite
  support; cf ergodic theorem for point
  processes~\cite[Theorem~13.4.III]{daley2007introduction}.
 For some other properties of the scattering moment estimators see~\cite[Section~5.1]{scatt}.

To avoid boundary effects and  enforce the translation invariance
of the empirical  scattering moments calculated over  finite rectangular window~$W$
it is customary to
map $W$ with the observerd points on the torus.

It is natural to restrict the  scale parameter to a finite window $j\in[j_{\min},j_{\max}]$.
The minimal scale  $j_{\min}$ is chosen such that 
the support of $\psi(j_{\min} ,\theta)$ separates points
(making $\hat S\tilde\phi(j_{\min },\theta)$
close to the empirical mean
$1/W\int_{W^2\times\R}u\tilde\theta\phi(\md(u,x))$
times the first norm of the mother wavelet). It does not depend on the angle $\theta$.
The  maximal scale  $j_{\max}$ is chosen such that $\psi(j_{\max},\theta)$
covers the whole window. {
For  the second order scattering moments, it is natural to
consider $j_{\min}+1<j_1<j_2\le j_{\max}$.
Regarding the choice of the angles in both families of scattering moments, it is  natural to consider
some common symmetric constellation of angles  $\theta^1,\ldots,\theta^a\in[0,2\pi)$.

Thus, the first and second  empirical  scattering moments 
calculated on $\tilde\phi$ form 
finite-dimensional (column) vectors.  We denote by $\hat\bfS\tilde\phi$,
and $\hat{\bfSS}\tilde\phi$, respectively,  the vector of
the first moments and   the  {\em joint} vector of the first {\em and}
second moments. If no ambiguity, for simplicity, in what follows we shall call
them just scattering moments of  $\tilde\phi$.

\subsection{Learning of marks via scattering moments}
\label{ss.Learnig-via-scattering}
Recall from Section~\ref{ss.Problem} that  our goal is
to learn the marking function $m$ (or $\tilde m$ depending on the 
auxiliary marks) from the training set of data, which consists of  examples  
of realizations of a marked point process $\Phi^m$
in a finite window, 
in order to calculate approximations of the unobserved  marks $m_i=m(x_i,\phi)$
for a new realization $\phi=\sum_i\delta_{x_i}$
of the (non-marked) point process $\Phi$.
Note that this problem consists  in the interpolation of
the function $m$ on the space $\R^2\times\mathbb{M}$ and due to the complexity
of this space   is not amenable to
any direct statistical approach. To overcome this difficulty,
we map this original problem to some finite dimensional regression problem and
solve it using  classical tools.

More specifically, in our main approach, 
we shall capture 
the function
$m$ through relation between the 
vector of the  first order scattering moments $\hat{\bfS}\phi^m$ of
the marked point pattern  
and the two moments  $\hat{\bfSS}\phi$
of the  non-marked one.  This relation is established using some
known regression models described in Section~\ref{sss.Regression}.
In order to be able to use it to
estimate marks  $m_i=m(x_i,\phi)$, we need next to
solve an inverse  problem described in
Section~\ref{sss.Reconstruction}. It consists in 
 reconstructing  marks from the
regressed scattering moments (approximating unknown
$\hat{\bfS}\phi^m$) knowing also  locations of points~$\phi$.

\subsubsection{Regression}
\label{sss.Regression}
Let $X_k:=\hat\bfSS\phi_k$ and  $Y_k:=\hat\bfS\phi^m_k$, $k=1,\ldots,n$,
be the vectors of the (empirical) scattering moments
calculated for the training data set consisting  of $n$ realizations  of the point process
$\Phi^m$, where  points 
$\phi_k=\sum_{i}\delta_{x_i(k)}$ and points with their marks 
$\phi_k^m=\sum_{i}\delta_{(x_i(k),m_i(k))}$ are considered,
respectively; cf Sections~\ref{ss.Problem} and \ref{sss.Scat-Estimate}.
Our goal is to find a common relation between
$X_k$ and $Y_k$ for all sample $k$,
and the simplest possible one is 
a linear relation represented by some matrix
$\mathbb{B}$ and vector $\boldsymbol\beta_0$
such that
\begin{equation}\label{e.Linear}
 \mathbb{B} X_k+\boldsymbol\beta_0 \approx Y_k\quad \text{for all $k=1,\ldots,n$.}
\end{equation}
This is a linear regression problem briefly presented 
in Section~\ref{par.Linear}. 
If the linear function does not allow one to capture the dependence,
we can use the  kernel regression or more advanced machine learning tools.
In this paper, we focus on the linear ridge regression.

\paragraph{Linear ridge regression}
\label{par.Linear}
To find the linear relation~\eqref{e.Linear}
we will use (linear) ridge model; cf~\cite[Section~7.5]{robert2014machine}.
 For $p=(j,\theta)$, denote by  $\boldsymbol{\beta}(p)$ the line of the matrix
$\mathbb{B}$ ~\eqref{e.Linear}; it
corresponds to the scattering moment in $Y_k$ at scale
$2^j$ and angle $\theta$, and similarly
the component $\beta_0(p)$
for the vector~$\boldsymbol\beta_0$.
Let  $Y_k(p):=\hat
S\phi^m_k(p)$ be the $(j,\theta)$-component  of $\hat\bfS^m_k$.
The ridge model consists in minimizing the regularized sum of the squared  residuals
$$\sum_{k=1}^{n} [ \boldsymbol{\beta}(p) X_k+\beta_0(p)  - Y_k(p)]^2+\lambda(p) ||\boldsymbol{\beta}(p)||^2\,,$$
for some  (Tikhonov) regularization parameter $\lambda(p)\ge 0$, chosen by the cross-validation (to minimize this squared  residuals on the validation set: a subset of the training set),
where $||\!\cdot\!||$ is the Euclidean norm.
This model, admits a well known  explicit solution in the form 
\begin{equation}\label{e.Linear-Ridge-Solution}
 [\hat{\beta}_0(p), \boldsymbol{\hat{\beta}}(p)]^{\mytop}
 =(\mathbb{X}^\mytop \mathbb{X}+\lambda(p)\mathbf{I})^{-1} \mathbb{X}^\mytop \mathbb{Y}(p),
\end{equation}
where  $\mathbb{X}$ is the matrix  with lines $X_k$ 
appended with the first column  of~1's, $\mathbb{Y}(p)$ is the column vector with elements
$Y_k(p)$, $k=1,\ldots,n$, $\mathbf{I}$ is the
appropriate identity matrix and~${}^\mytop$ stands for the matrix
transpose.

Using~\eqref{e.Linear-Ridge-Solution}  one can calculate approximations
$\hat{S}\phi^m(p)$ of the scattering moments of a new  marked
configuration $\phi^m=\sum_i\delta_{(x_i,m_i)}$
observing only its points  $\phi=\sum_i\delta_{x_i}$
\begin{align}\label{e.Linear-Ridge-Estimation}
 \doublehat{S}\phi^m(p)&:=\boldsymbol{\hat {\beta} }(p)\,\hat\bfSS\phi+ \hat{ \beta} _0(p)
   \nonumber
\end{align}
where $\hat\bfSS\phi$ is the vector of  the 2nd order scattering moments calculated
on~$\phi$  (points-only).
Remember, expression~\eqref{e.Linear-Ridge-Solution}  requires tuning of
the regularization parameters $\lambda(p)\ge 0$ usually needed
in high dimensional regression problems when the matrix $\mathbb{X}^\mytop \mathbb{X}$ is not invertible.
The ordinary least square  (OLS) estimator corresponding to 
$\lambda(j,\theta)=0$ is not performing well  in this case.



%

\subsubsection{Reconstruction}
\label{sss.Reconstruction}
Using linear ridge  we
calculate approximations $\doublehat{S}\phi^m(j,\theta)$ 
of the scattering moments of a new  marked
configuration $\phi^m=\sum_i\delta_{(x_i,m_i)}$
observing only its points  $\phi=\sum_i\delta_{x_i}$.
Denote  the whole vector of $\doublehat{S}\phi^m(j,\theta)$ by $\doublehat{\bfS}\phi^m$. 
From $\doublehat{\bfS}\phi^m$ we estimate (reconstruct) unknown  marks
$m_i$ of $\phi^m$
looking for a solution to  the following minimization problem
\begin{equation}\label{e.reconstruction}
\arg\min_{\phi^{'m}:\phi'=\phi}||\hat\bfS\phi^{'m}- \doublehat{\bfS}\phi^m||^2,
\end{equation}
where we minimize over all arbitrarily marked configurations $\phi^{'m}$
sharing the points with given $\phi$ (hence  over
unknown  marks) and  $\hat\bfS\phi^{'m}$ denotes the scattering moment
calculated  for  $\phi^{'m}$.
It should be noted that~\eqref{e.reconstruction}  is a non convex
optimization problem.  To solve it
we use  L-BFGS-B algorithm, which 
is  a limited-memory algorithm for solving large nonlinear optimization problems
subject to simple bounds on the variables. It is intended for problems in which information on
the Hessian matrix is difficult to obtain, or for large dense problems; cf~\cite{lbfgs,lbfgs2}.

\subsection{Learning via local distance representation}
\label{ss.Learnig-benchmark}
In this approach, considered as a benchmark, we consider each 
marked point $(x_i(k),m_i(k))$ of each realization
$\phi^m_k$, $k=1,\ldots,n$, of the training set; cf
Section~\ref{par.Linear},  {\em along with some neighborhood},  as one
element  of  the  new training data set. In this case the
training set  consists of point  patterns having a  marked
point the  center,  surrounded by  some  neighboring  points. Using
the the linear  ridge regression
described in Section~\ref{par.Linear}, mutatis mutandis,
we regress the marks of those central points
with respect to the vectors containing all inter-point distances in the considered
point neighborhood, ordering the points  according to the distance to
the central point and flattening the distance matrix in (say) raw
major order. 
Note that, in this approach, there is no reconstruction phase, as the
marks are directly approximated.
The main parameter of this approach is the number of points~$K$ taken into
account in the neighborhood.

\section{Numerical results}
\label{s.Results}
In this section we provide the details of our   numerical study of the
main problem of this paper, namely  statistical learning of geometric
marks of point processes.
We begin by describing some general assumptions and procedures.

\subsection{General numerical framework} 

\subsubsection{Scattering moment approach}
For our numerical experiments, for each specific geometric model,
we create a data set (denote it by  $\mathcal{X}$) of Poisson point
patterns $\phi_k$ (with constant intensity to be specified) 
using the R software and its package {\em Spatstat} for point process
analysis~\cite{baddeley2005spatstat}.
The points are considered in the unit  square window and their
marks are analytically computed (according to the given model) using {\em Spatstat} with the
window  mapped to the torus, thus leading to the marked point patterns
$\phi^m_k$, $k=1,\ldots,n$. The size of the data set $\mathcal{X}$
is $n=10\,000$ marked point patterns.

The  (empirical) scattering moments are computed on these point patterns
 (with and without marks) using {\em ScatNet}
software~\cite{anden2014scatnet} developed in Matlab
(no {\em Spatstat} implementation is available yet).
It uses  a zero-mean variant of the  Morlet wavelet; cf Example~\ref{example:Morlet}.
This latter software  working on raster images, we convert each marked point
pattern of $\mathcal{X}$ into images of size $2^7\times2^7$ pixels 
(removing images with points corresponding to the same pixel).

The following family of scattering moments are computed.
We assume  the smallest scale $j_{\min}=0$. At this scale the first
scattering moments correspond
  to the empirical  mean measure and do  not depend on the
  angle $\theta$; cf  Section~\ref{sss.Scat-Estimate}.
We take $j_{\max}=7$ and  following constellation of 8 angles  $ \theta = 0,
  \pi/8,\ldots, 7\pi/8$ for all scales $j=1,\ldots,7$.
Thus, there are $1 + 8 \times 7 = 57$  first order scattering 
moments (dimension of the regressed vectors $\hat\bfS_k^m$ equals to 57)
and $8\times8\times\frac{7\times6}{2} =1344$ 
of the moments of the second order (thus making the dimension of the
explanatory vectors  $\hat\bfSS_k$   equal to $57+1344=1401$).

We use the   linear ridge  regression  described in Section~\ref{sss.Regression}
on the data set $\mathcal{X}$.
To optimize the regression parameters we make 5-fold
cross-validation~\cite[Section~1.4.8]{robert2014machine} on
$\mathcal{X}$.

Having calculated the  estimators $\doublehat\bfS^m\phi$  of the  first marked scattering moments
$\hat\bfS^m\phi$ for the point patterns in the test set
 we use L-BFGS-B algorithm~\cite{lbfgs,lbfgs2}
to solve the inverse problem~\eqref{e.reconstruction}, that is to
reconstruct the marks.
This is a  steepest descent algorithm for which it is important to
optimize (via cross-validation) the
number of iterations; to be explained  in
Section~\ref{par.Reconstruction-discussion}.
Using  a rule of thumb (no formal cross-validation) 
we fix the number of steps so as to minimize the mean square
error (MSE) on the test data set.

\subsubsection{Benchamrk}
As explained in Section~\ref{ss.Learnig-benchmark},
for every image in the data set $\mathcal{X}$, we consider each point
of the image, along with its  $K$ neighbours  as an element of a new
data set $\mathcal{X'}$. More precisely, we take the first $20,000$ of
points of the images of $\mathcal{X}$ and create 
$\mathcal{X'}$ with  $K$ neighbours of the chosen  point.
The value of $K$ depends on range of dependence of a given mark model;
it is experimentally discovered for each mark, as will be explained in
Section~\ref{par.Choice-K}. 
We use $\mathcal{X}'$ to regress directly,  using the linear  ridge
regression method, the (one dimensional)  central point marks
with respect to the vector  of dimension $K(K-1)$
of the local distance matrix. There is no reconstruction phase in the benchmark.

\begin{figure}[t]
\vspace{-2ex}
  \begin{center}
\includegraphics[width=1\linewidth,height=0.25\linewidth]{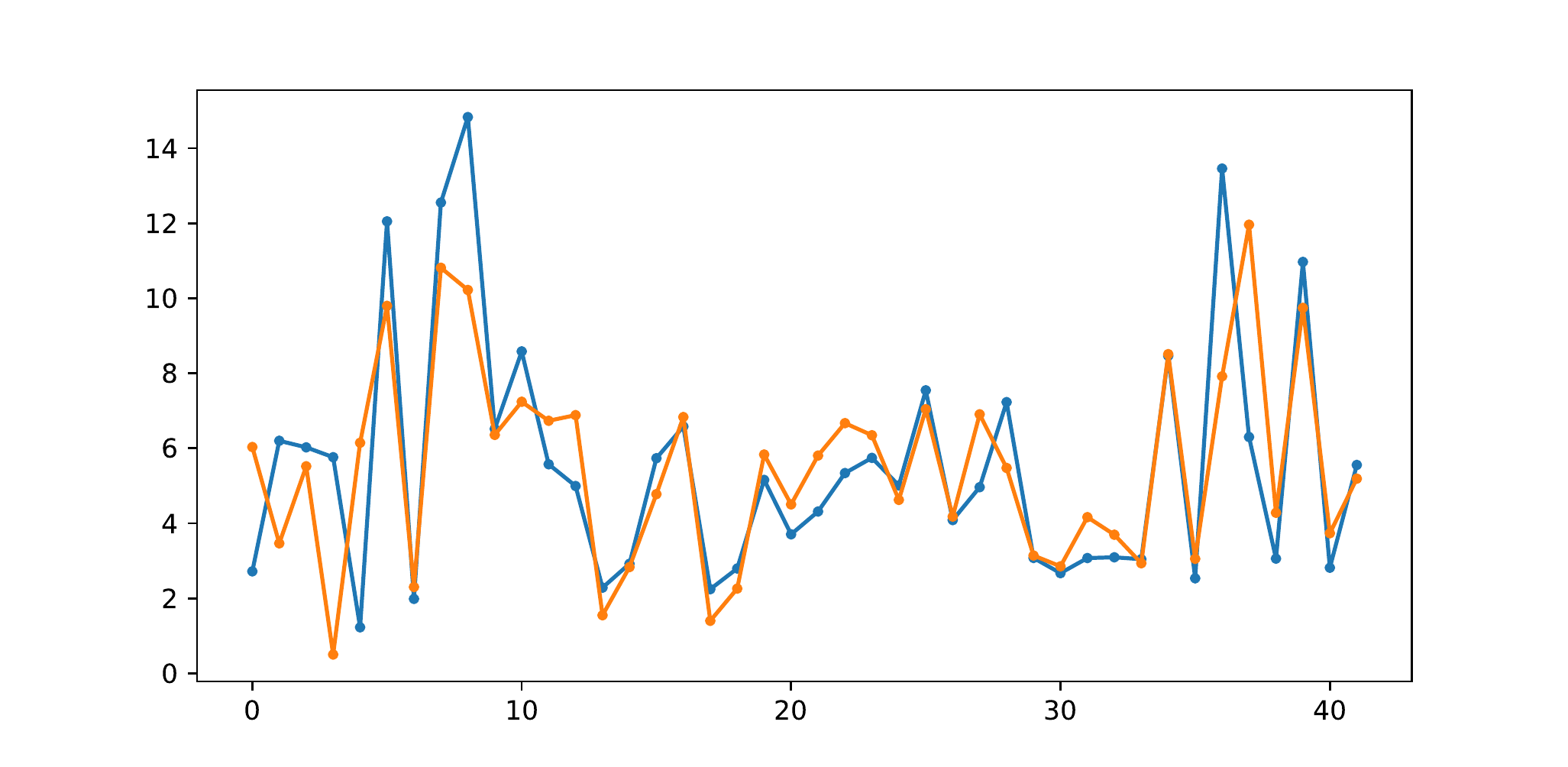}\\[-1ex]
\caption*{\footnotesize Reconstructed image  example (mapped to
                  2D; the peaks of the lines correspond to the  values of marks of points 
   numbered  in lexicographic order; orange curve --- reconstruction,
   blue --- exact values). }
\vspace{-1ex}
\end{center}
%
\vspace{-2ex}
  \begin{center}
\hbox{\includegraphics[width=0.35\linewidth]{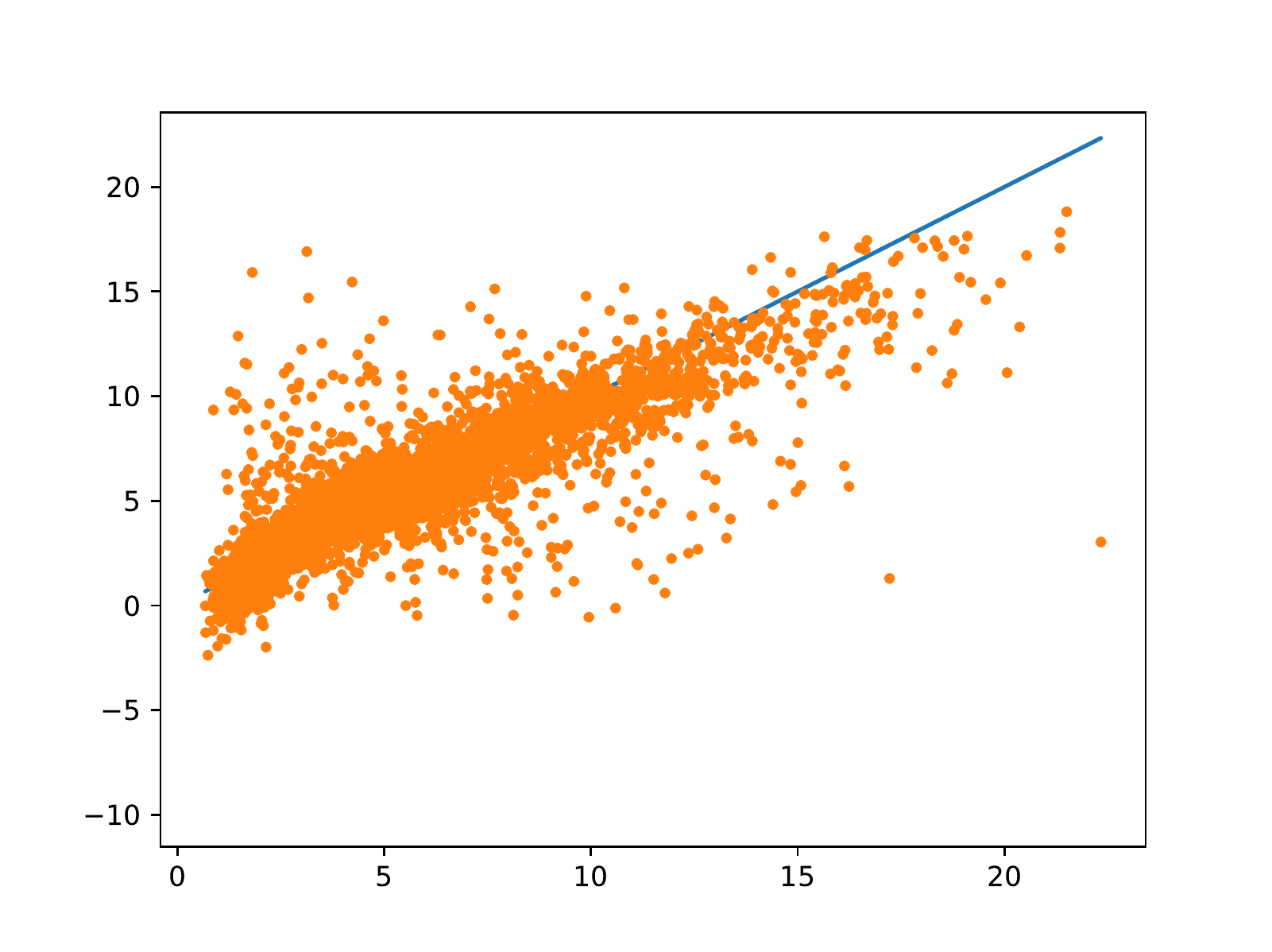}\hspace{-1em}
    \includegraphics[width=0.35\linewidth]{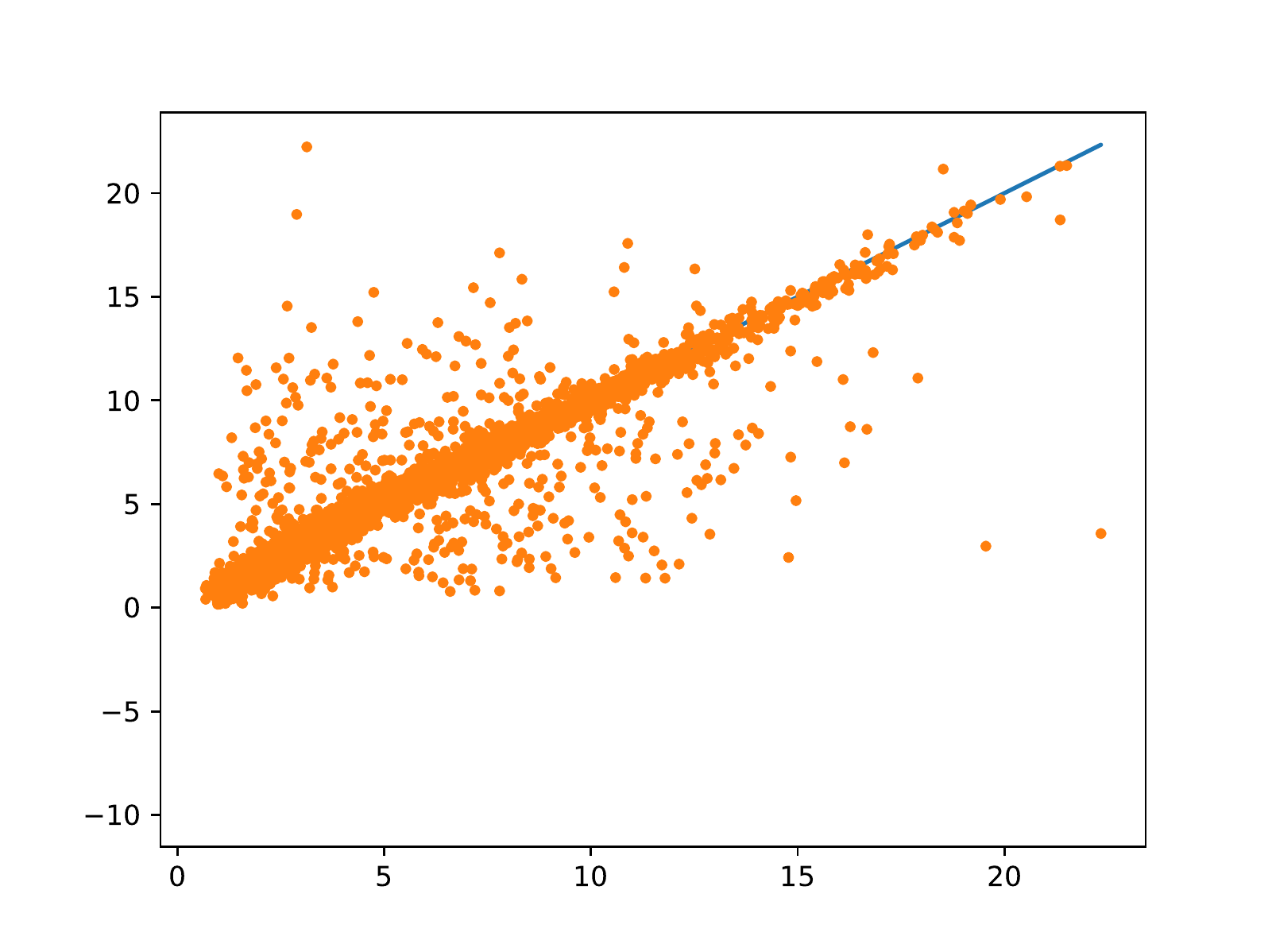}\hspace{-1em}
    \includegraphics[width=0.35\linewidth]{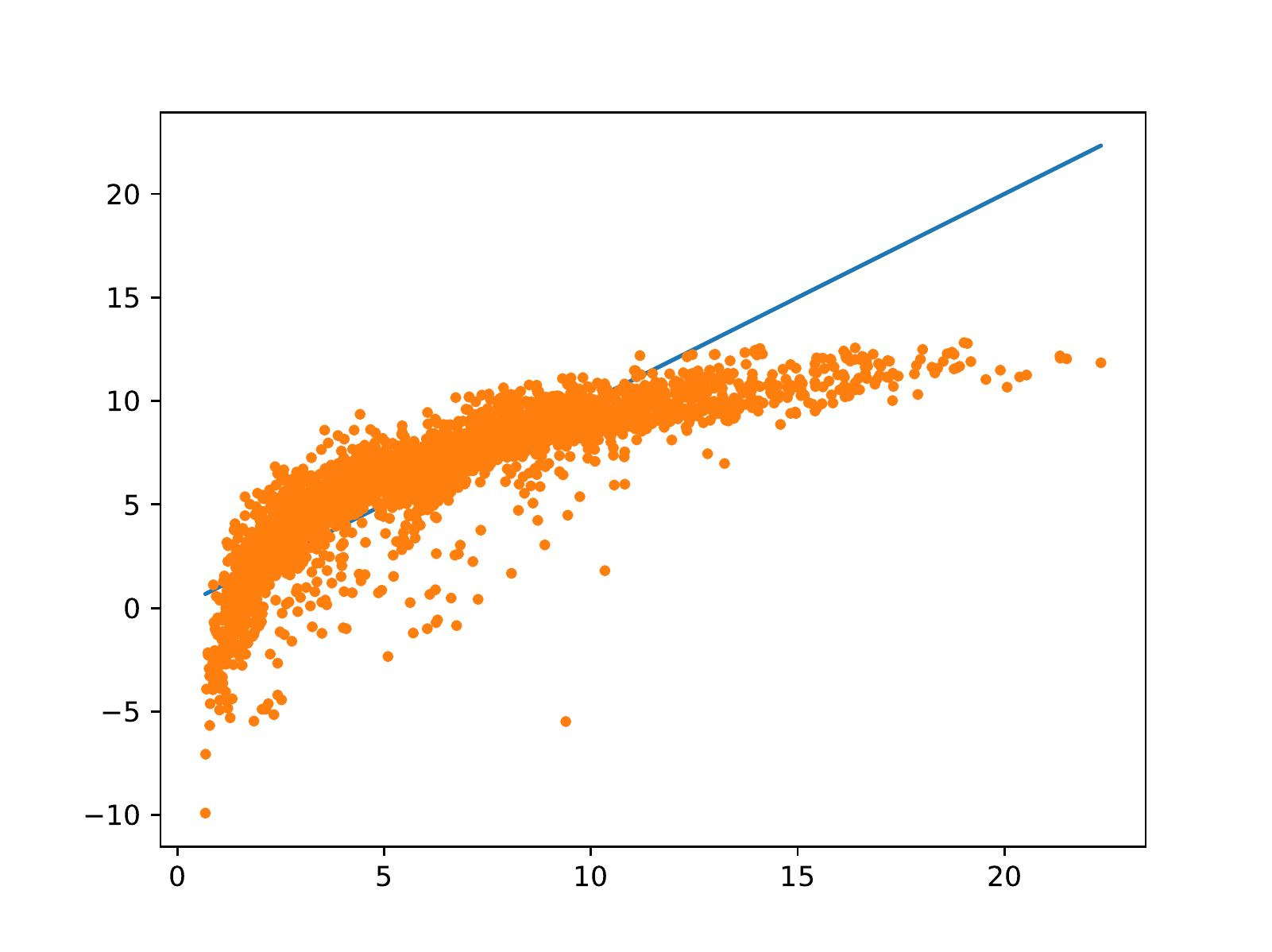}}
\vspace{-2ex}
\centerline{\footnotesize (a)\hspace{0.3\linewidth}(b)\hspace{0.3\linewidth}(c)}
\caption{Shot-noise reconstruction. Up: an example of image
  reconstruction using scattering moments. Down:  Q-Q plots (reconstructed mark in function
    of its true value) for  100 test images (a) regressed scattering
    moments,  (b) exact scattering moments, (c) benchmark.}
\label{fig.SN-cloud}
  \end{center}
  \vspace{-4ex}
\end{figure}
\begin{table}[t]
\begin{center}
{\small
\begin{tabular}{|l||l|l|l|l|}
  \hline
  method &scattering & scattering &
  exact& bench-\\[-1ex]
  &\scriptsize $5\,000$ & 
  \scriptsize $10\,000$&scatt.&\ \ \ mark\\
   \hline\hline
   RMSE       & 1.99 & 1.98 &
                                                            1.64 & 1.98 \\
   \hline
   NRMSE1 & 9.21e-2 & 9.15e-2 &
                                                                        7.56e-2 & 9.15e-2 \\
   \hline
   NRMSE2 & 3.32e-1 & 3.29e-1 & 
                                                                    2.75e-1 & 3.29e-1 \\
 \hline
\end{tabular}}
\vspace{-1ex}
\caption{Shot-noise   reconstruction errors  using different methods. 
\label{shot_comp}}
\end{center}
\vspace{-4ex}
\end{table}

\begin{figure}[t]
  \vspace{-2ex}
\begin{center}
\includegraphics[width=1\linewidth, height=0.25\linewidth]{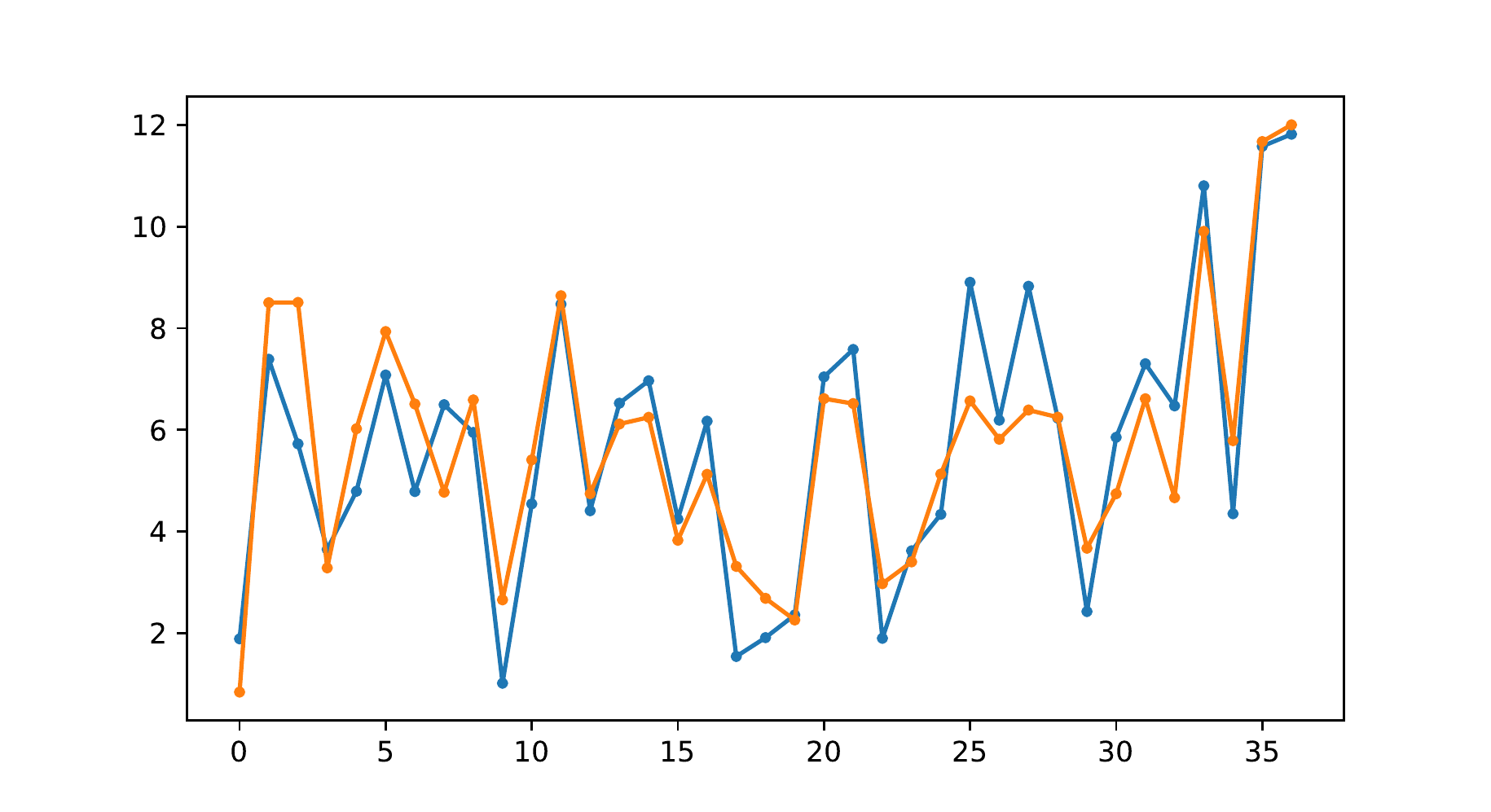}\\[-1ex]
\caption*{\footnotesize Reconstructed image  example.}
\includegraphics[width=0.35\linewidth]{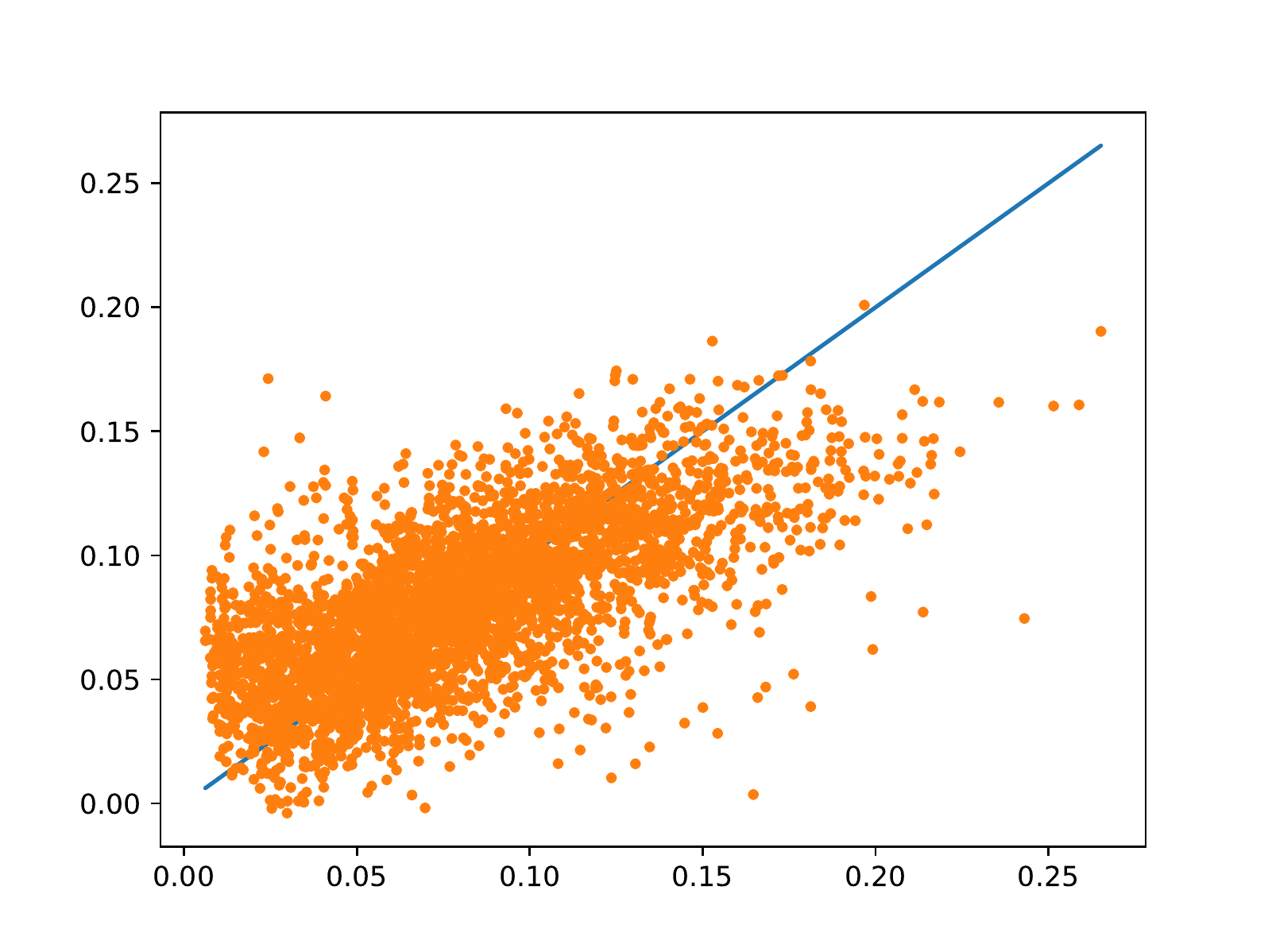}
     \includegraphics[width=0.35\linewidth]{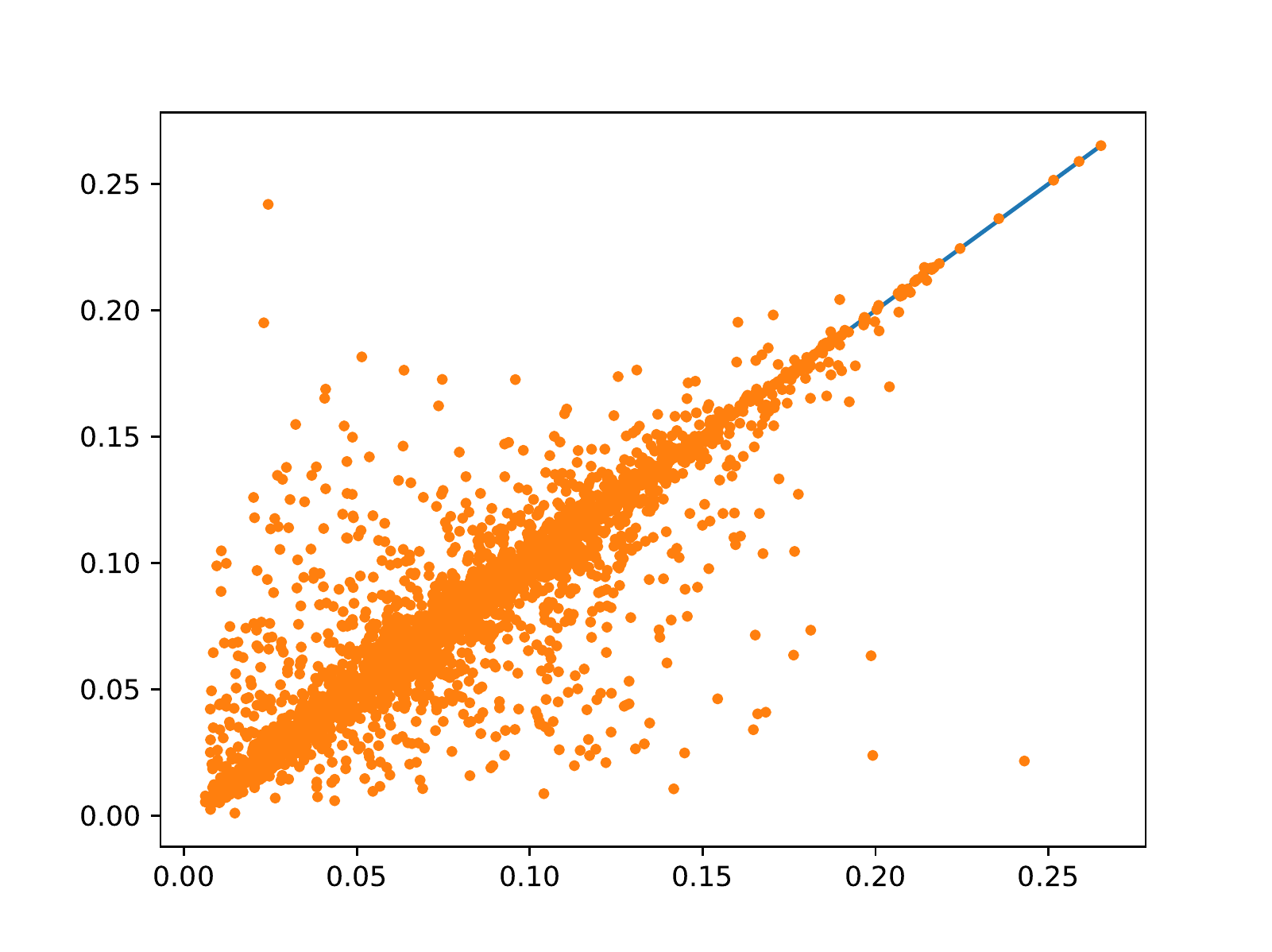}\\[-2ex]
     \centerline{\footnotesize(a)\hspace{0.3\linewidth}(b)}
  \caption{Nearest neighbour distance reconstruction; example and Q-Q
  plots using: (a) estimated scattering moments, (b) exact scattering moments.}
\label{fig.NN-cloud}
 \end{center}
 \vspace{-3ex}
 \end{figure}
\begin{table}[t]
    \begin{center}
{\small \begin{tabular}{|l||l|l|l|l|}
  \hline
method& estimated scatt. & exact scatt. & benchmark\\
\hline\hline
RMSE & 3.14e-2 & 
                                              2.06e-2 & -- \\
   \hline
   NRMSE1 & 1.21e-1 &
                                                       7.94e-2 & -- \\
   \hline
   NRMSE2  & 3.94e-1& 
                                                       2.58e-1 & -- \\
   \hline
\end{tabular}}
\caption{Nearest neighbour distance; reconstruction error}
\label{nn_comp}
\end{center}
\vspace{-5ex}
\end{table}

\subsubsection{Validation methodology}
\label{sss.Validation}
For testing  of the main (scattering moment) approach,
we produce an independent data set of 100 marked
point pattern realizations and we use the first 100 points of this set 
for the benchmark approach.

For all points in the respective test sets, 
we compute Q-Q plots and the root mean square error (RMSE), the normalized RMSE
with normalization by the  range ($\max - \min$) of the marks (NRMSE1),
and the normalized RMSE with normalization by the mean of the marks
(NRMSE2).

In order to study the quality of the
representation of the mark point patterns  by their  first order
scattering moments (independently of the regression), we
also perform  the reconstruction of marks~\ref{sss.Reconstruction}
directly from the exact  (and not regressed)
scattering moments~$\hat\bfS^m\phi$. Note, these latter  are not ``exact'' marks
but marks reconstructed from exact empirical scattering moments of the given
point pattern. The gap between them and  the marks reconstructed from
the regressed moments allows one to apprehend the error introduced by
the regression.

\subsection{Results}
\label{ss.GM-results}

We now present our numerical study  of different mark models
presented in Section~\ref{s.Marks-model}.
The observed results are discussed in Section~\ref{sss.Discussion}.
\subsubsection{Shot-noise}
\label{ss.SN-results}

We consider shot-noise marks introduced in Section~\ref{sss.SN-model}
with the response function 
$\ell(r):=\max(10r,0.6)^{-3}$.
For this example the  data set $\mathcal{X}$ consists of $10\,000$
realizations of  Poisson point process of intensity 40
(recall, always considered the  unit  square window).
In order to observe how the performance of our approach
depends on the size of the data set and on two different regression
methods,  we first use only $5\,000$ elements of $\mathcal{X}$ as the
training set for the  linear ridge regression to estimate the scattering
moments of the marked point patterns. Next, we use the whole training
set, with linear ridge regression. The neighborhood for the benchmark
is $K=15$. The number of iterations for the reconstruction from the
exact scattering moments  and estimated ones is taken, respectively,
30 and 4; see Section~\ref{par.Reconstruction-discussion} for the explanation.

Figure~\ref{fig.SN-cloud} presents an example of the reconstruction of
marks of one given image by the main method and Q-Q plots for various
reconstruction variants. Table~\ref{shot_comp} presents reconstruction
errors. 
Recall the  Q-Q plot (c) and 
the last column of the table represents the error of the
reconstruction of the marks from the exact (and not regressed)
scattering moments.

\subsubsection{Nearest neighbour distance}

The data set for this example is also made of 10000 realizations of a
Poisson point process with intensity 40. In this case, in
On Figure~\ref{fig.NN-cloud} and in Table~\ref{nn_comp}
we only show the results for the entire training set.
For this mark model we do not consider  the benchmark, because
the nearest neighbour distance is equal to the 
 first element of the distance matrix. Thus, the scattering moment
 approach cannot do better.
 The number of iterations for the reconstruction from the
exact scattering moments  and estimated ones is taken, respectively,
250 and  8.

\subsubsection{Voronoi cell surface area}
\label{sss.Voronoi-area-results}
 \begin{figure}[t]
   \vspace{-2ex}
   \begin{center}
\includegraphics[width=1\linewidth, height=0.25\linewidth]{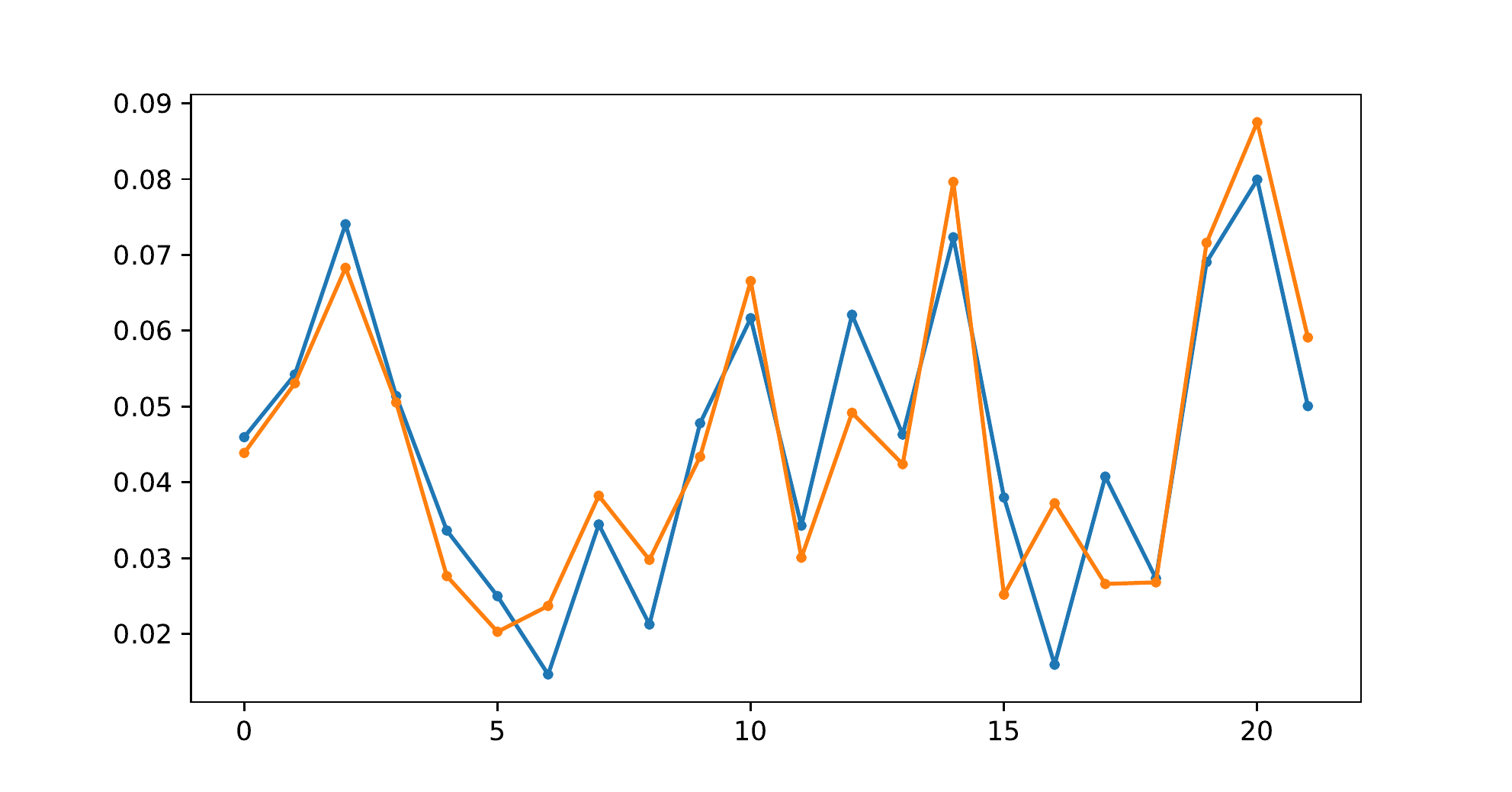}\\[-1ex]
\caption*{\footnotesize Reconstructed image  example.}
\includegraphics[width=0.35\linewidth]{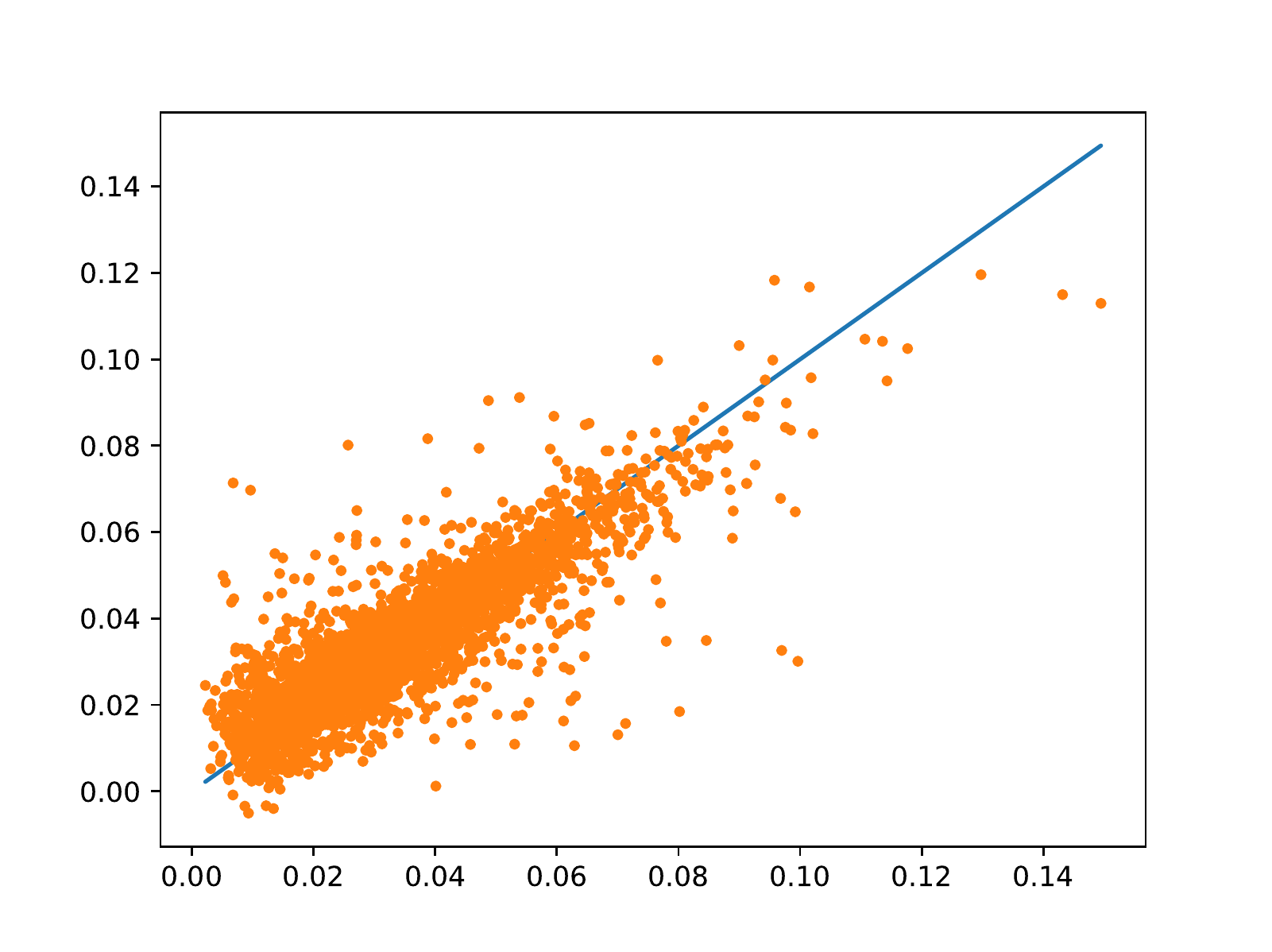}\hspace{-1em}
\includegraphics[width=0.35\linewidth]{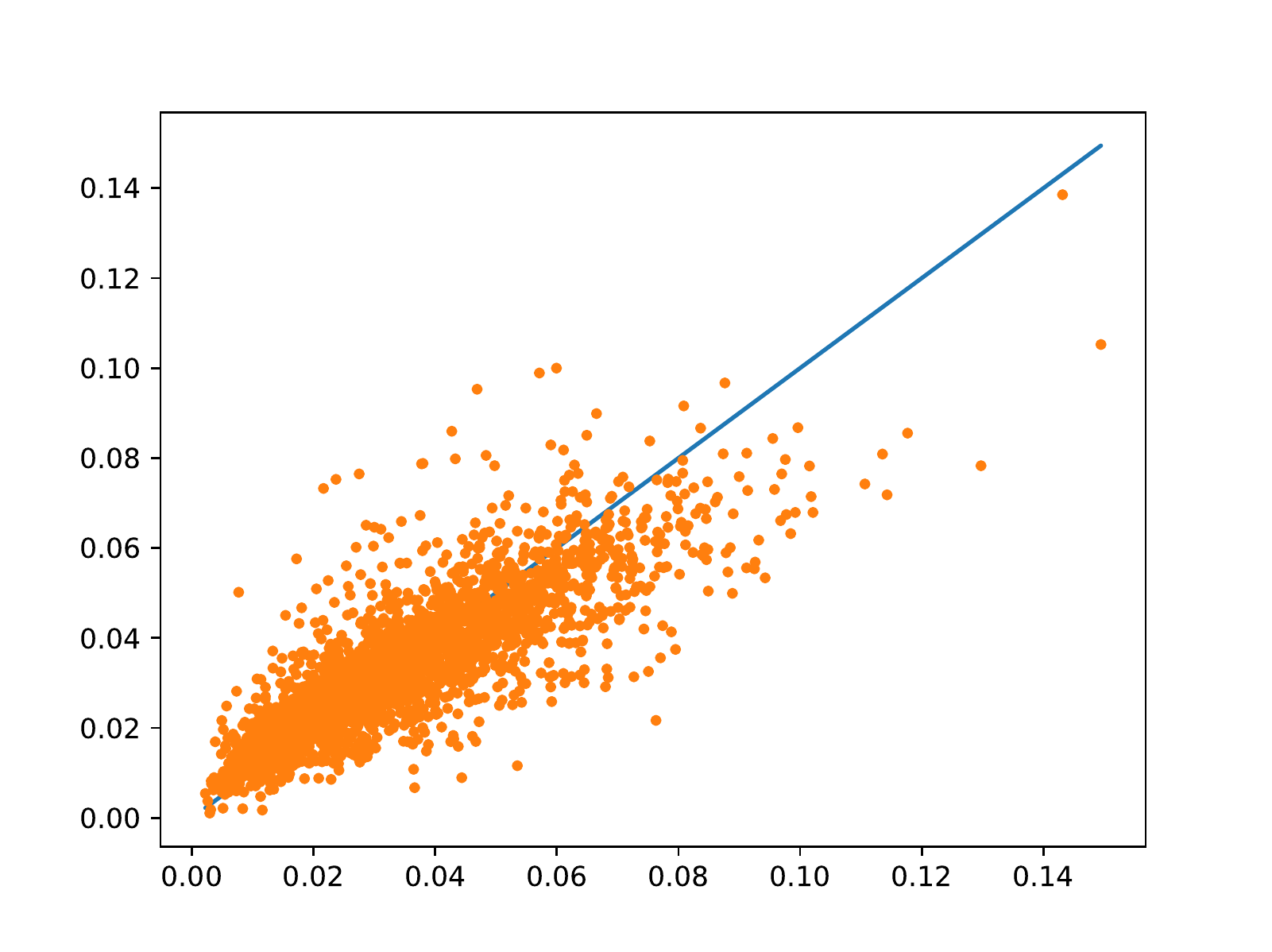}\hspace{-1em}
\includegraphics[width=0.35\linewidth]{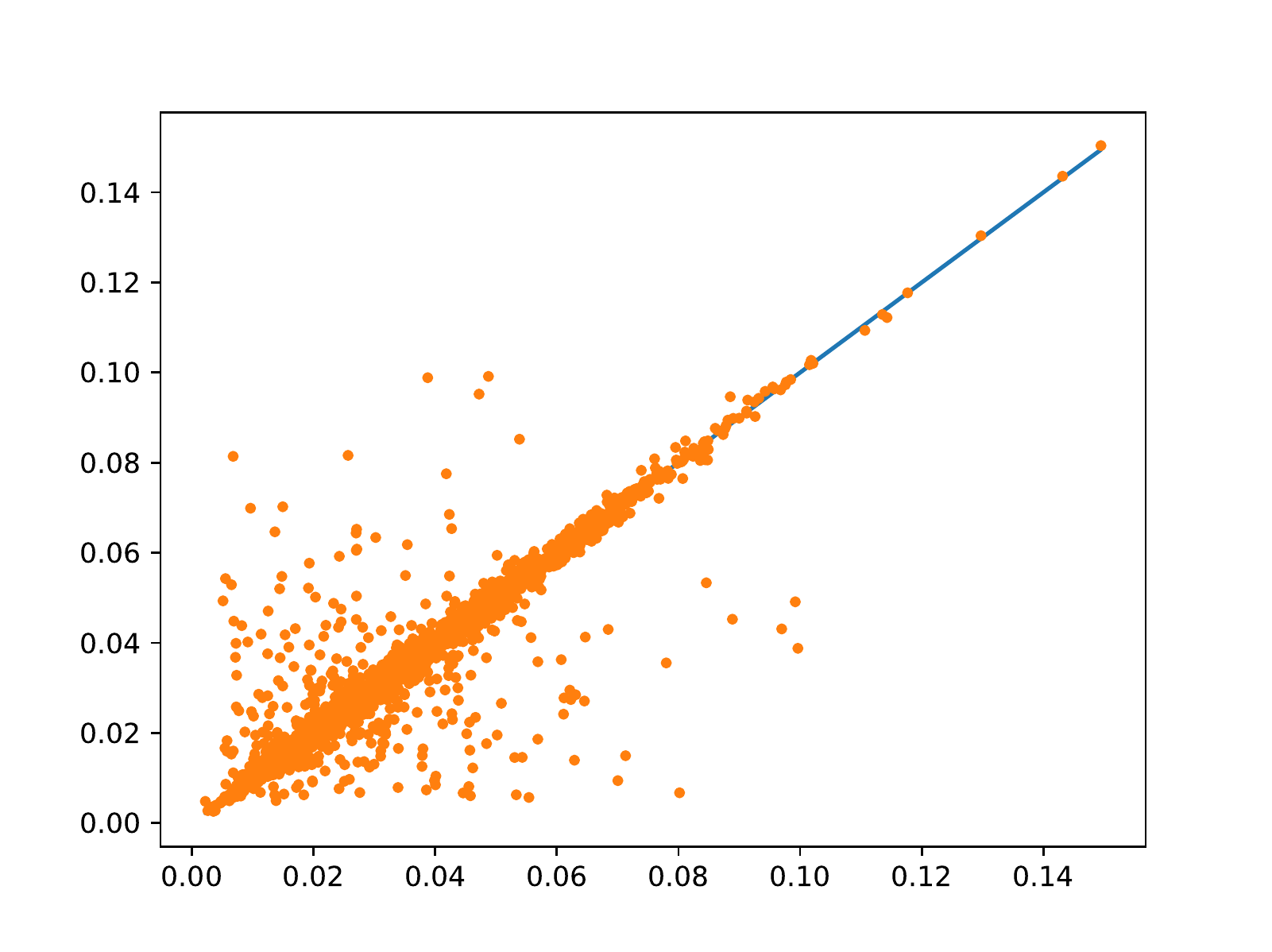}\\[-2ex]
\centerline{\footnotesize (a)\hspace{0.3\linewidth}(b)\hspace{0.3\linewidth}(c)}
\caption{Voronoi cell surface area reconstruction; example and Q-Q
  plots using: (a) estimated scattering moments,  (b) exact scattering
  moments, (c)  benchmark.}
\label{fig.Voronoi-cloud}
 \end{center}
 \vspace{-3ex}
\end{figure}
\begin{table}[t]
\begin{center}
{\small \begin{tabular}{|l||l|l|l|l|}
 \hline
method& estimated scatt.& exact scatt.  & benchmark \\
\hline\hline
   RMSE       & 9.71e-3  & 7.33e-3 & 9.61e-2 \\
   \hline
   NRMSE1 & 6.60e-2 & 4.98e-2  & 6.53e-2 \\
   \hline
   NRMSE2  & 2.87e-1  & 2.16e-1 & 2.85e-1 \\
   \hline
\end{tabular}}
\caption{Voronoi cell surface  reconstruction errors.}
\label{vor_comp}
\end{center}
\vspace{-5ex}
\end{table}

For this example, the training set $\mathcal X$ consists of 10\,000
realizations of  Poisson point process with intensity 30.
The reconstruction results  are presented on Figure~\ref{fig.Voronoi-cloud}
and in Table~\ref{vor_comp}.
The neighborhood for the benchmark
is very large $K=35$ (more than the average number of points)
because the sum of the areas of the
Voronoi cells in the finite window 
is constant equal to the total window surface area, which introduces a
strong global dependence
for this mark.
The number of iterations for the reconstruction from the
  exact scattering moments  and estimated ones is taken, respectively,
  30 and 6.

\subsubsection{Voronoi cells moment of inertia}
\label{sss.Voronoi-inertia-results}
For this example, similarly to the previous example, the training set
$\mathcal X$ consists of 10\,000 realizations of a Poisson point process
with intensity 30.
The results are presented on Figure~\ref{fig.Inertia-cloud} and 
in Table~\ref{mom_comp}.
 The neighborhood for the benchmark
 is $K=15$ (note that the global dependence specific for the Voronoi
 surface area does exist here). 
 The number of iterations for the reconstruction from the
exact scattering moments  and estimated ones is taken, respectively,
150 and 8.


\begin{figure}[t]
\vspace{-2ex}
  \begin{center}
\includegraphics[width=1\linewidth,height=0.25\linewidth]{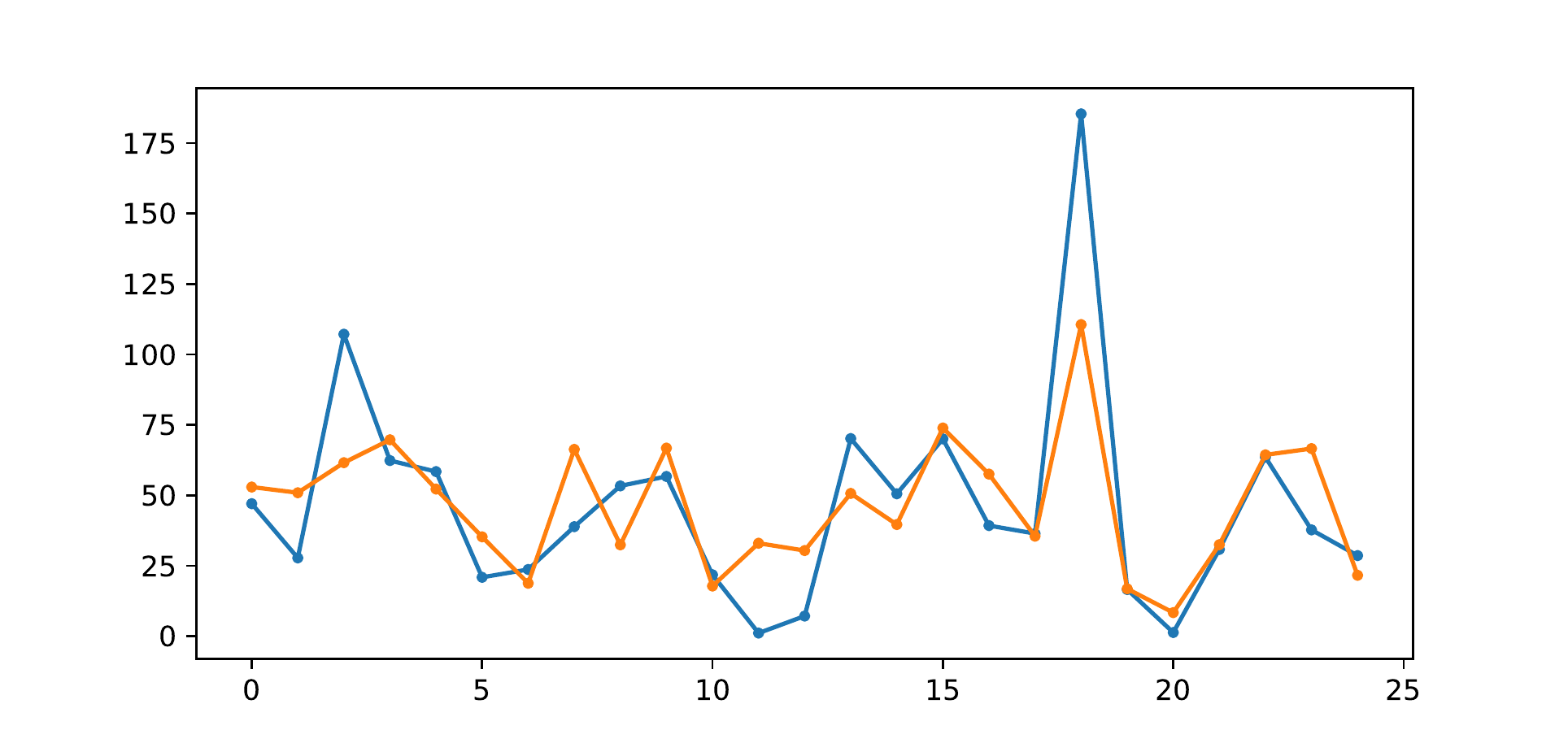}\\[-1ex]
\caption*{\footnotesize Reconstructed image  example.}
\hbox{\includegraphics[width=0.35\linewidth]{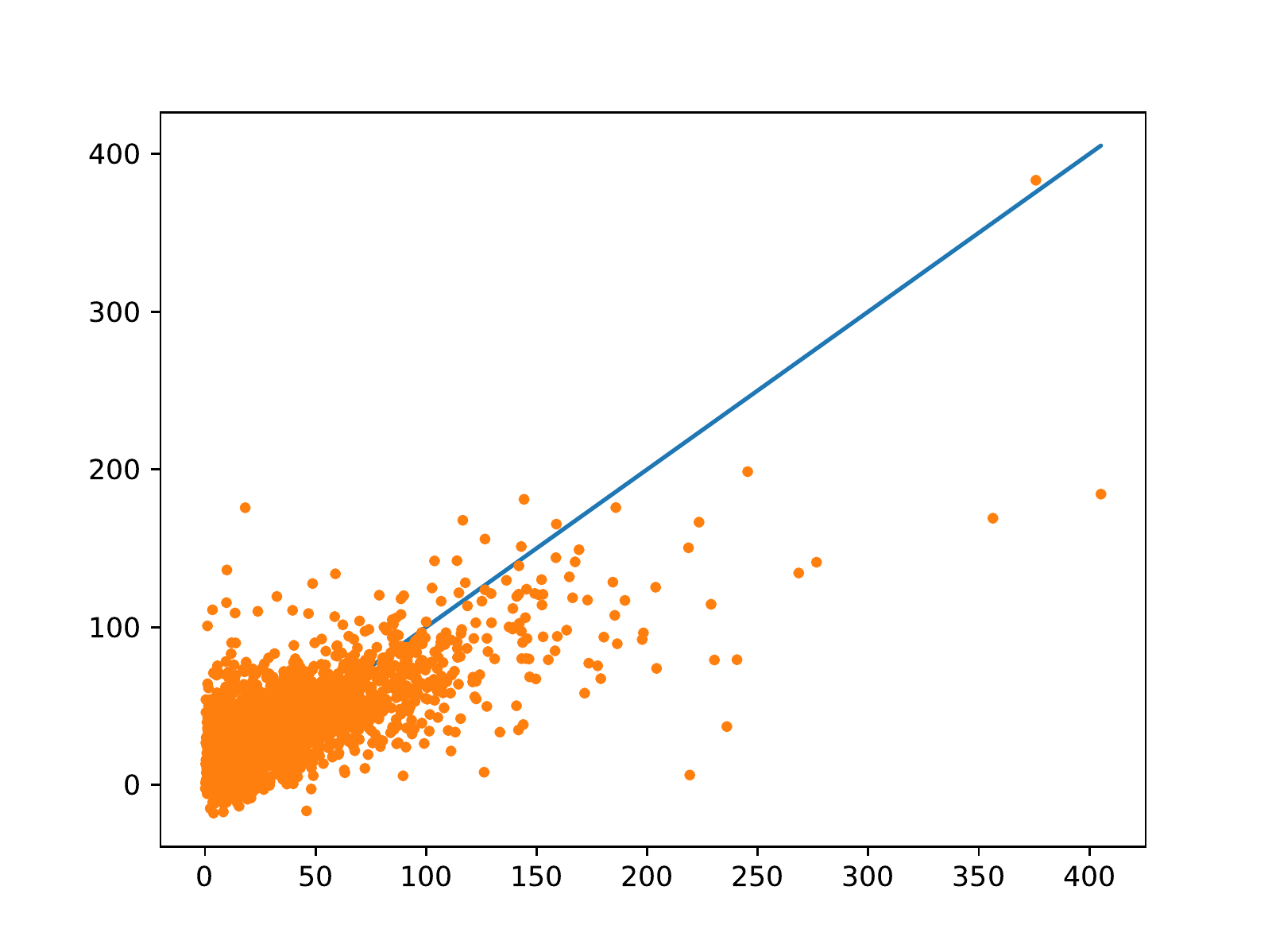}\hspace{-1em}
    \includegraphics[width=0.35\linewidth]{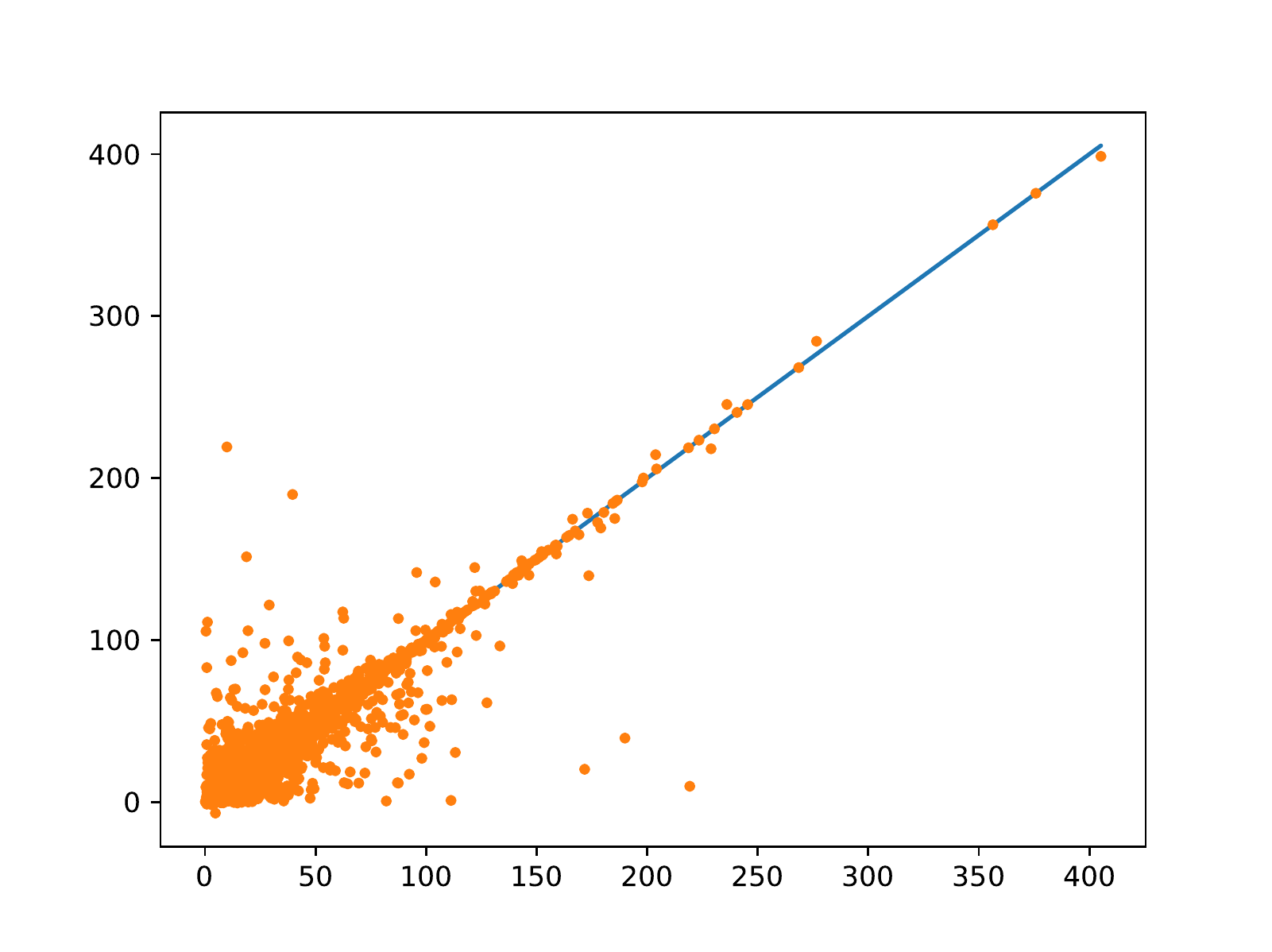}\hspace{-1em}
    \includegraphics[width=0.35\linewidth]{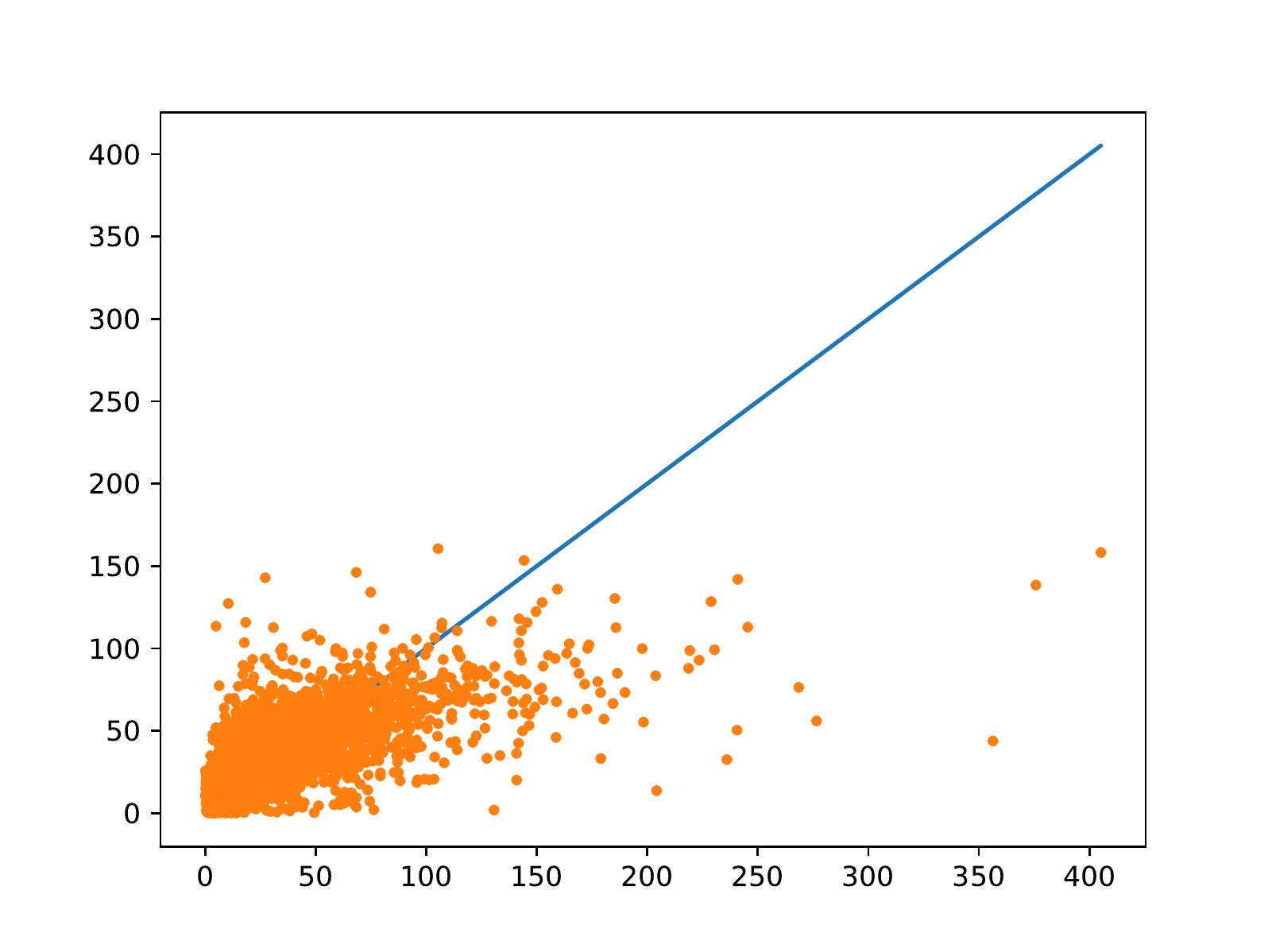}}
\vspace{-2ex}
\centerline{\footnotesize (a)\hspace{0.3\linewidth}(b)\hspace{0.3\linewidth}(c)}
 \caption{Voronoi cell moment of inertia reconstruction; example and
   Q-Q points using (a) estimated  scattering  moments, (b)  exact
   scattering moments, (c) benchmark.}
 \label{fig.Inertia-cloud}
 \end{center}
 \vspace{-3ex}
\end{figure}
 \begin{table}[t]
\begin{center}
{\small\begin{tabular}{|l||l|l|l|}
  \hline 
method&   estimated scatt.& exact scatt. &benchmark\\
   \hline   \hline
   RMSE        & 2.42e-4 & 1.37e-4 & 2.65e-4 \\
   \hline
   NRMSE1  & 5.98-2 & 3.39e-2  & 6.53e-2 \\
   \hline
   NRMSE2   & 7.12e-1 & 4.04e-1 & 7.74e-1 \\
   \hline
\end{tabular}}
\caption{Voronoi cell moment of inertia  reconstruction errors.}
\label{mom_comp}
\end{center}
\vspace{-5ex}
 \end{table}

\subsubsection{Voronoi shot-noise}
\label{sss.Voronoi-sn-results}
For this example, similarly to the previous example, the training set
$\mathcal X$ consists of 10\,000 realizations of a Poisson point process
with intensity 30.
The results are presented on Figure~\ref{fig.VSN-cloud} and 
in Table~\ref{tab.VSN}.
 The neighborhood for the benchmark
 is $K=15$ . 
 The number of iterations for the reconstruction from the
exact scattering moments  and estimated ones is taken, respectively,
50 and 5.


 \begin{figure}[t]
   \vspace{-2ex}
 \begin{center}
\includegraphics[width=1\linewidth,height=0.25\linewidth]{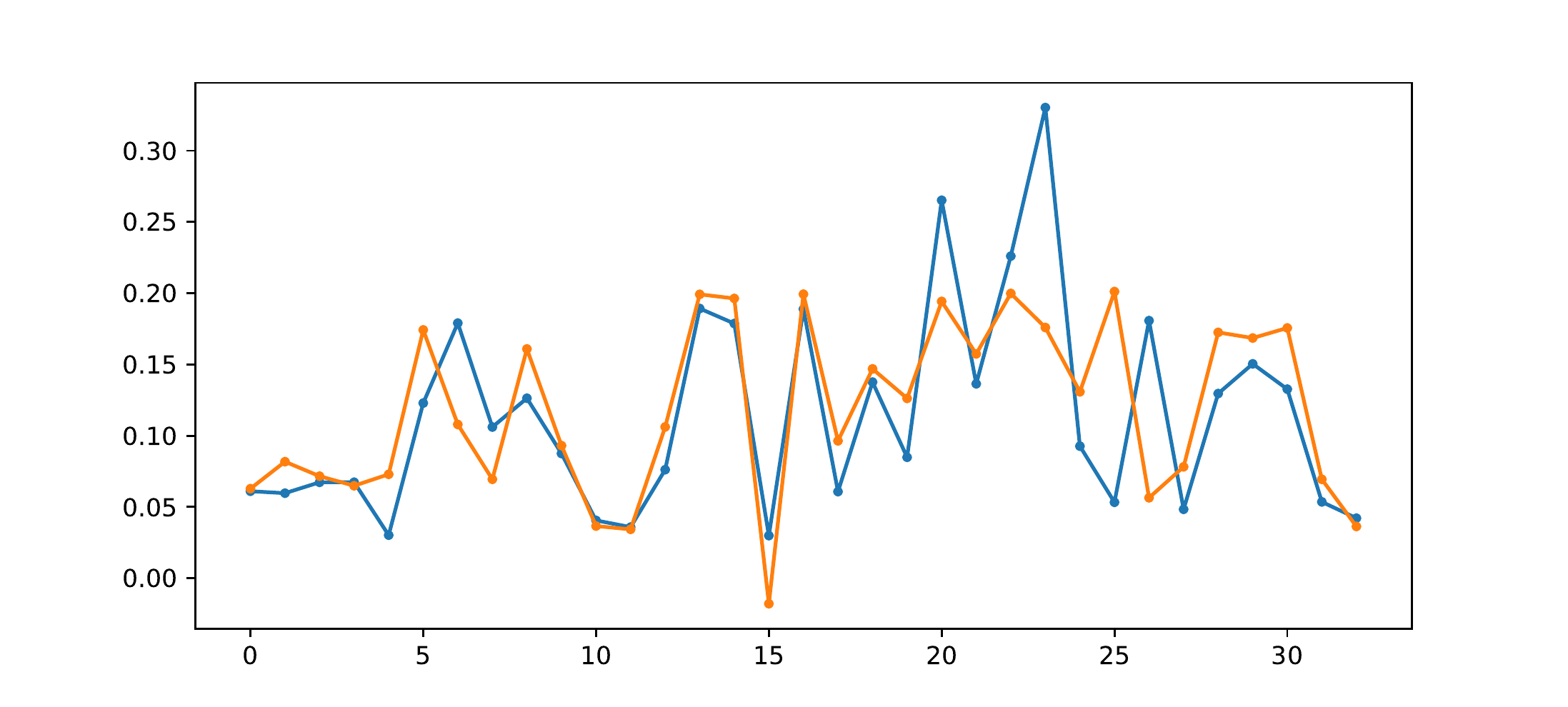}\\[-1ex]
\caption*{\footnotesize Reconstructed image  example.}
\hbox{\includegraphics[width=0.35\linewidth]{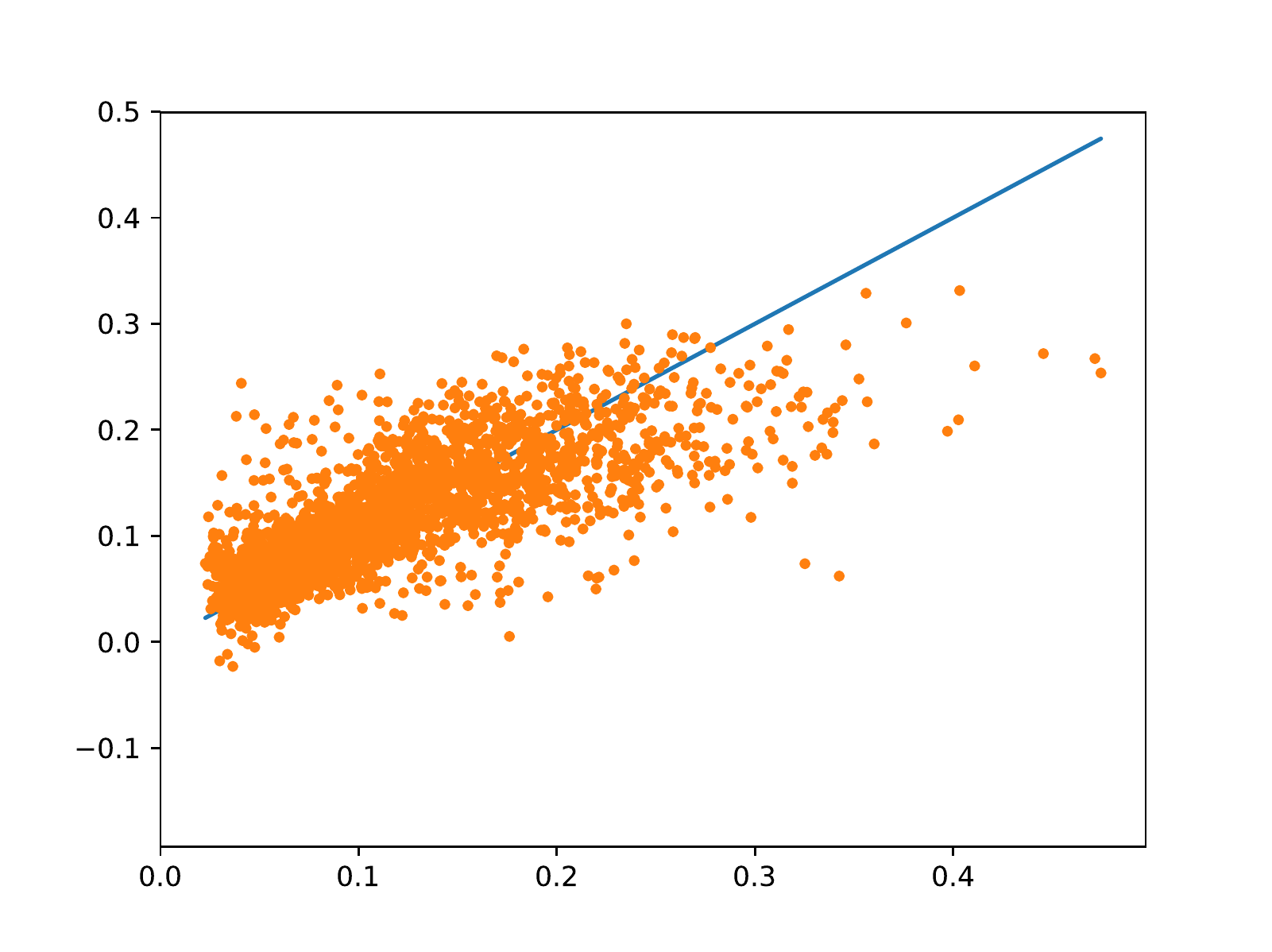}\hspace{-1em}
    \includegraphics[width=0.35\linewidth]{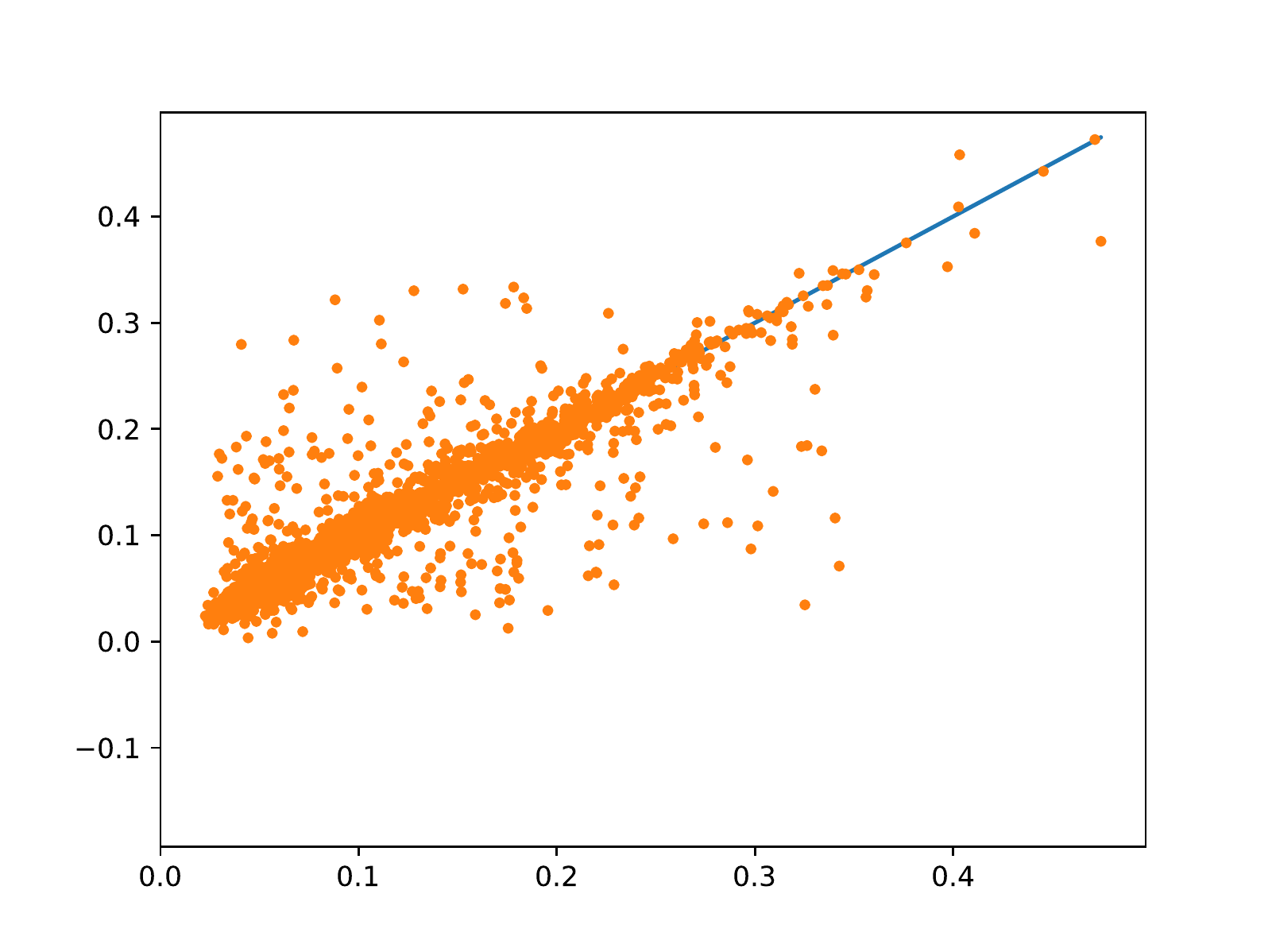}\hspace{-1em}
    \includegraphics[width=0.35\linewidth]{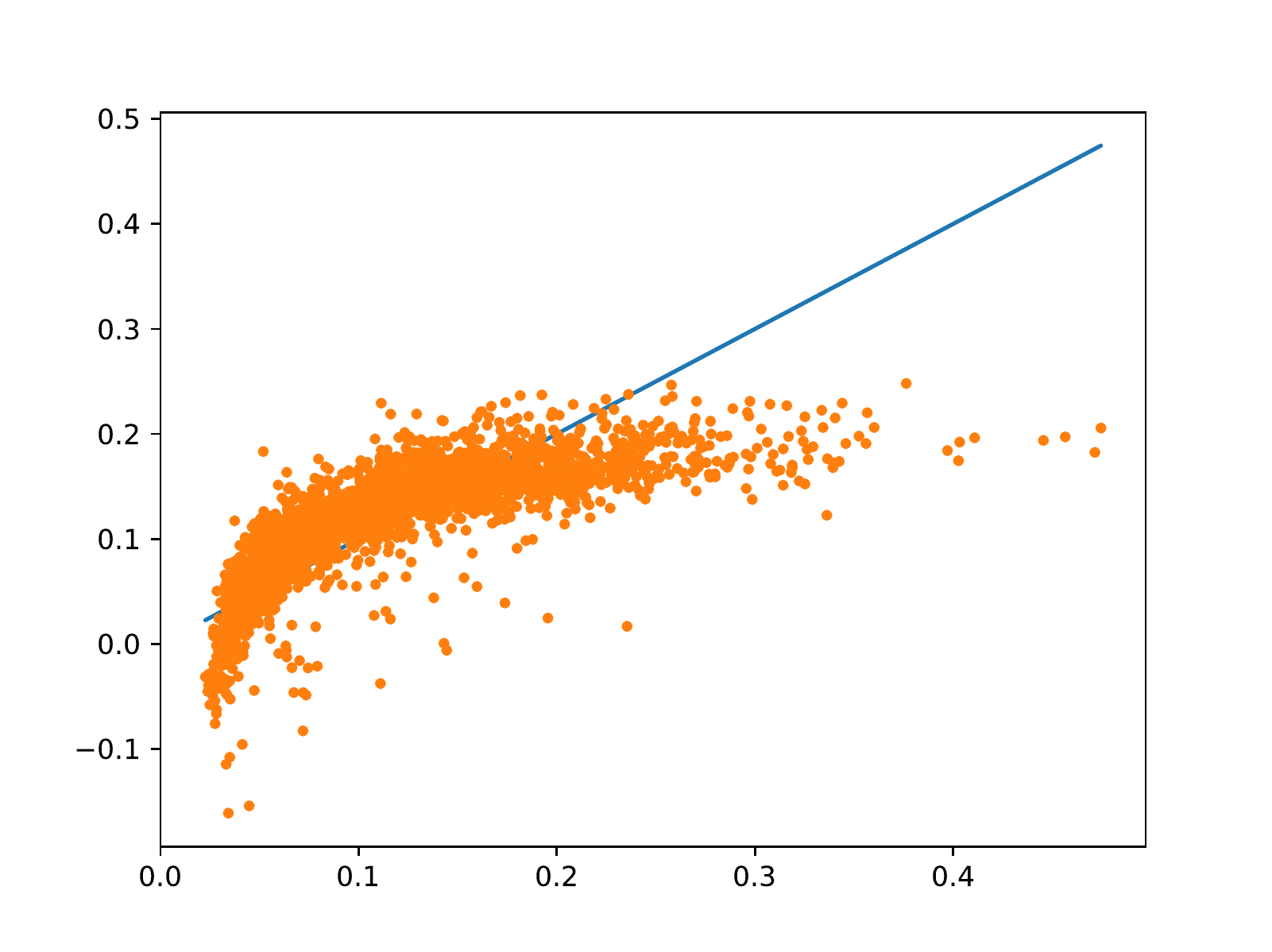}}
\vspace{-2ex}
\centerline{\footnotesize (a)\hspace{0.3\linewidth}(b)\hspace{0.3\linewidth}(c)}
 \caption{Voronoi shot-noise reconstruction; example and
   Q-Q points using (a) estimated  scattering  moments, (b)  exact
   scattering moments, (c) benchmark.}
\label{fig.VSN-cloud}
 \end{center}
 \vspace{-3ex}
\end{figure}
\begin{table}[t]
\begin{center}
{\small \begin{tabular}{|l||l|l|l|}
  \hline 
method&   estimated scatt.& exact scatt. &benchmark\\
   \hline   \hline
  RMSE        & 4.17e-2 & 3.02e-2 & 4.29e-2 \\
   \hline
   NRMSE1  & 9.24-2 &  6.69e-2 &  9.50e-2\\
   \hline
   NRMSE2   & 3.76e-1 & 2.72e-1 &  3.86e-1\\
     \hline
\end{tabular}}
\caption{Voronoi shot-noise  reconstruction errors.}
\label{tab.VSN}
\end{center}
\vspace{-5ex}
\end{table}

\begin{figure}[t]
  \vspace{-2ex}
\begin{center}
\includegraphics[width=1\linewidth,height=0.25\linewidth]{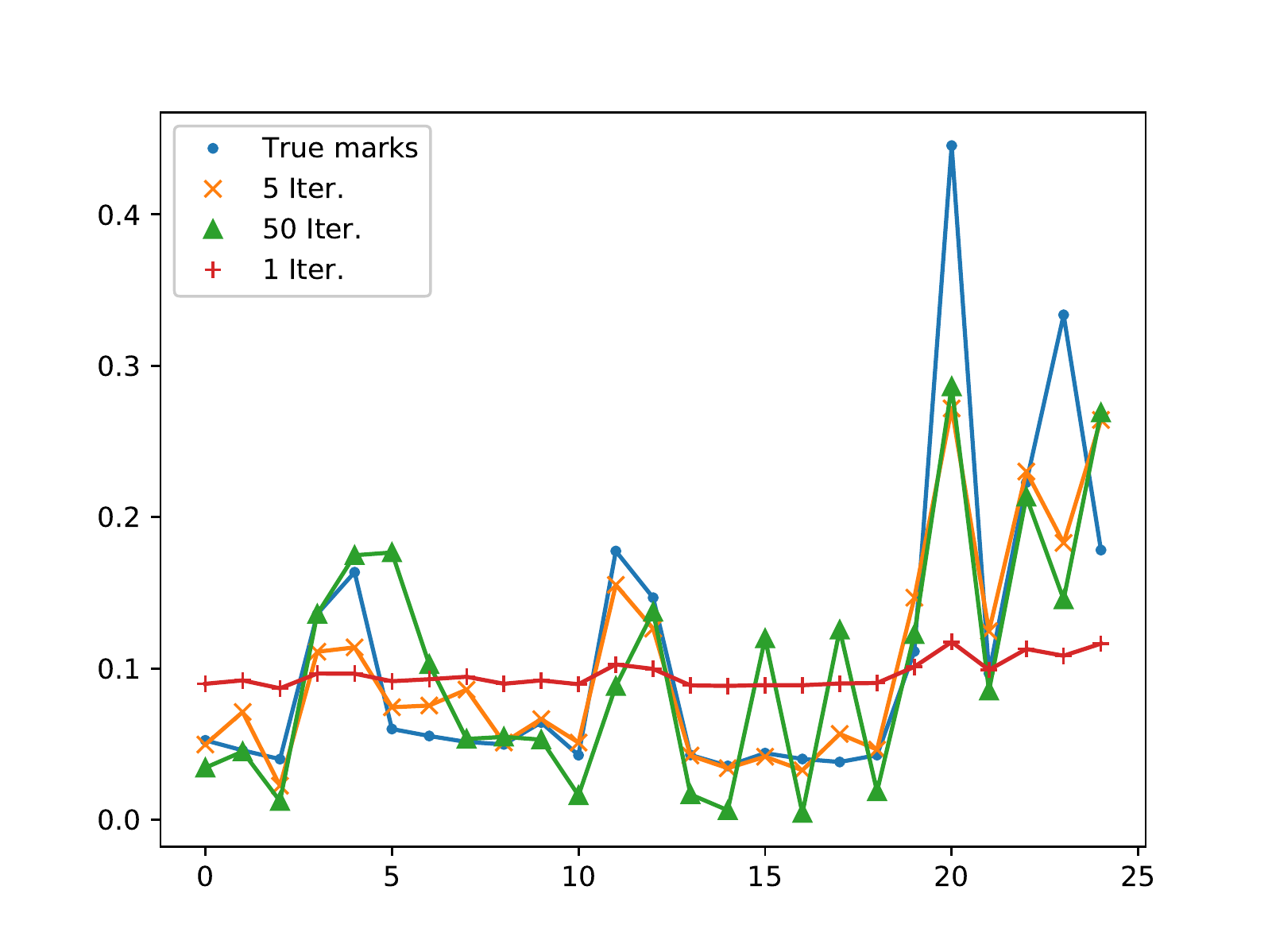}\\[-2ex]
  \caption{Reconstruction of a sample of the Voronoi shot-noise marks from
    the estimated scattering moments after 1, 5 and 50 iterations. 5
    iterations give the smallest RMSE of 4.17e-2, while  50 iterations
    used for the reconstruction from the exact scattering moments
    improve the reconstruction of some marks but give
    worse RMSE of 5.26e-2.
   \label{fig.reconstruction-iterations}}
\end{center}
\vspace{-2ex}
  \vspace{-2ex}
\begin{center}
 \includegraphics[width=1\linewidth,height=0.25\linewidth]{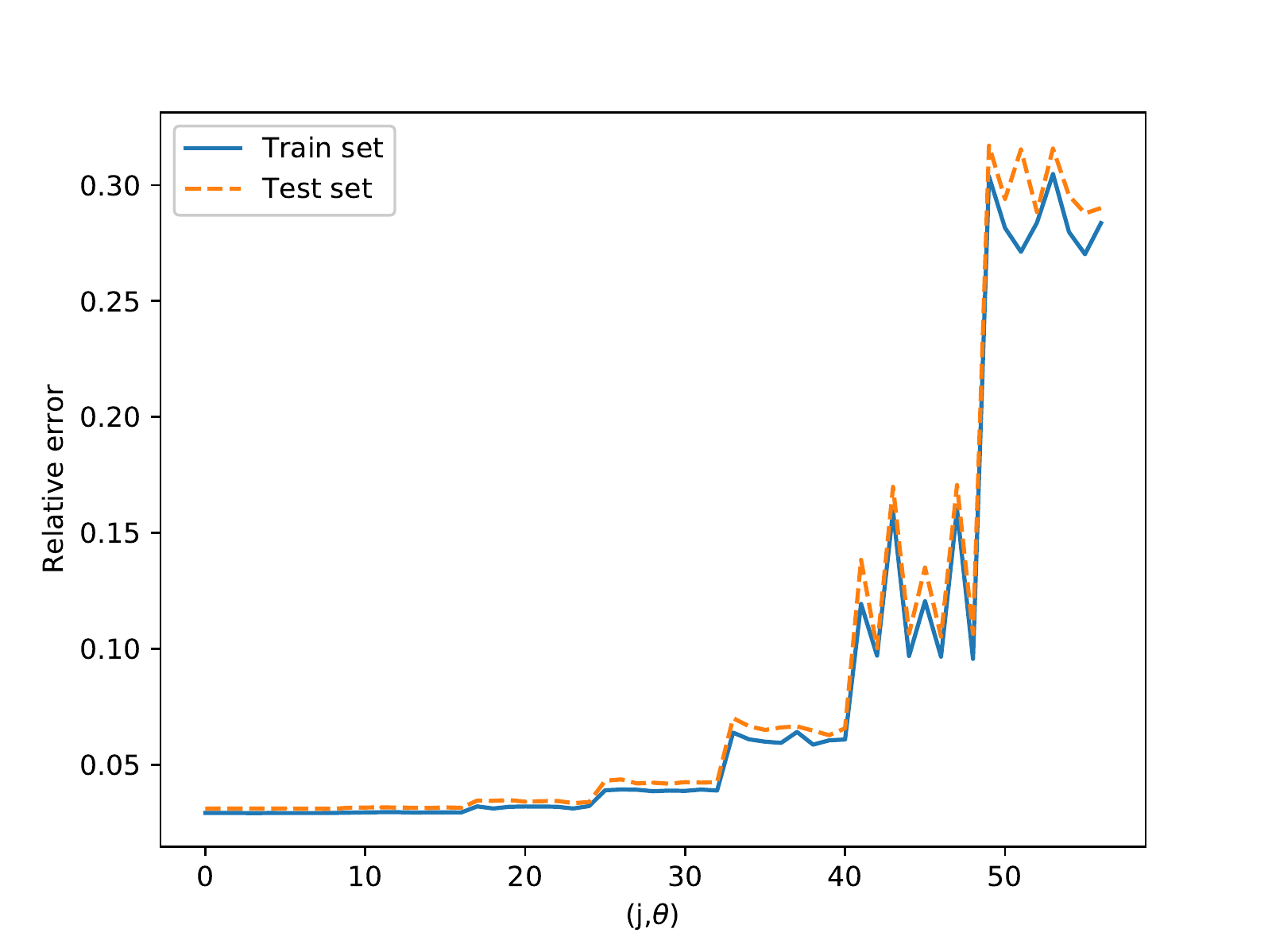}
 \caption{Regression relative errors $\epsilon^m(j,\theta)$ for $(j, \theta)$ in order
   $(0,0), (1,\pi/8),\ldots, (2,0), \ldots,(7,7\pi/8)$; calculated on
   training and test set.
         \label{fig.regression-rel-errs}}
\end{center}
\vspace{-2ex}
\end{figure}

\begin{table}[t]
\begin{center}
  {\small 
  \begin{tabular}{|l||l|l|l|l|}
    \hline
  model\hfill $K$
          &  10 & 15 & 20 & 35\\
   \hline\hline
 Shot-noise   & 1.99 & 1.98 & 1.98 & --\\
   \hline
   Voronoi area & 1.04e-2 &  1.00e-2 &  9.80e-3 & 9.61e-3\\
   \hline
   Voronoi inertia  & 2.76e-4 & 2.65e-4 &  2.80e-4 & --\\
   \hline
   Voronoi shot-noise  & 4.32e-2  & 4.30e-2 &  4.29e-2 & --\\
   \hline
\end{tabular}}
\caption{Choice of the number of neighbours $K$ for the benchmark
  approach and the corresponding RMSE.}
\label{choice_k}
\end{center}
\vspace{-5ex}
\end{table}

\subsection{Discussion}
\label{sss.Discussion}
The following remarks can be formulated regarding the observed results.
\subsubsection{Reconstruction from the exact and estimated  moments}
\label{par.Reconstruction-discussion}
Observing the Q-Q plots and (N)RMSE's  of the  marks reconstructed from
the exact first order scattering moments, we see ``how much
information'' they effectively  carry  regarding the marking function.
While all marks are relatively well  represented in this way,
the quality of the reconstruction depends on the type of dependence
represented by a given mark. For example the shot-noise and the
surface areas of the Voronoi cell are more easy to represent
than the nearest neighbour.
This can be explained by different  sensitivity (stability) to small
deformations of the point pattern, with a precise formulation yet to be
theoretically  studied on the ground of point processes.

A typical, significant error  in the image reconstruction both from the exact end the estimated scattering moments  consist in the swap of a large and
a small  mark of two neighbouring points (e.g. the points number 36 and 37 on
e.g.~Figure~\ref{fig.SN-cloud}) not  leading to a significant
modification of the considered  scattering moments.
This effect can be seen also on the Q-Q plots where many points significantly far from the diagonal appear in symmetric pairs.
We believe these swapping errors should occur less often  for more regular (less clustering) point processes than the considered Poisson one. Examples of such point processes are determinantal ones; see~\cite{dcx-clust,blaszczyszyn2015clustering} for the clustering comparison theory. Future studies should investigate this issue.

As we have already mentioned,  it is
important to properly tune the number of iterations of the
steepest descent algorithm used in the reconstruction phase, preventing it from going too deeply into
potential local minima. We observe the following  {\em local-global reconstruction quality 
trade-off}:  While some number of  initial  iterations makes  all the marks approach
their right values, further iterations improve the quality of approximation of
some subset of marks at the price of degrading  this quality for the
remaining ones; cf Figure~\ref{fig.reconstruction-iterations}.
This trade-off can  be observed on average by watching the  RMSE  on the
test set, which first decreases and then increases. 
We use this observation to choose an optimal number of  iterations.
It is larger for the reconstruction from the exact scattering moments
than for the reconstruction from the estimated ones, where there is
less incentive to force the algorithm to approach the values of the
scattering moment values  that  themselves are not exact.
This additional reduction of the quality of reconstruction from
the estimated moments is particularly penalizing the scattering
learning approach, where similar or even better quality of the
regression might not be good   enough as the input for the reconstruction phase.

 \subsubsection{Choice of neighbourhood}
\label{par.Choice-K}
A crucial  benchmark parameter is the number of neighbors to be
considered in the local representation. They should be selected in function of
the type of mark dependence. If no a priori information is available,
this can be done observing the RMSE on the test set as shown in Table~\ref{choice_k}.
Observe that the constant sum of the Voronoi cell areas makes them
globally dependent, unlike the  Voronoi moments of inertia.

\subsubsection{Quality of  the regression}
The regression relative errors  calculated 
 $\epsilon^m(j,\phi):=|\doublehat{S}\phi^m(j,\theta)-\hat{S}\phi^m(j,\theta)|/\hat{S}\phi^m(j,\theta)$
calculated on the training and test set,  presented on
Figure~\ref{fig.regression-rel-errs}, show that there is no
overfitting in the regression of the scattering moments.

\subsubsection{Scattering moments versus benchmark}
Regarding the RMSE scattering  outperforms the benchmark for all marks
except for the Voronoi surface area with the neighbouhood taking
almost all points and,  for obvious reasons, the nearest neighbor
distance. A spectacular difference can be observed on the Q-Q plots of the
shot-noise on Figure~\ref{fig.SN-cloud} and the Voronoi shot-noise on
figure~\ref{fig.VSN-cloud} where the benchmark 
essentially fails to capture the marking function.
This shows that the benchmark approach might not be appropriate in the
case of long-range dependent marks.

\section{Concluding remarks}

Motivated by the stochastic-geometry problems related to wireless networks, in this paper we have discussed how to learn the point-marking function dependent on the configurations of points. 
We propose two different approaches to address the problem using the ideas from statistical learning.
The baseline approach extracts the geometric information for each
point based on the matrix distance of its  nearby points. It is then
solved using the ridge linear regression method. The difficulty is to
choose the number of nearby points. 
The other approach uses the (multiscale wavelet) scattering moments to define a global feature vector for all the points in the domain, and another feature vector for the points with marks. 
The relation between the two  feature vectors is also learned using linear regression. 
Then we use an image super-resolution technique~\cite{bruna2015super} 
to reconstruct the marks  from the predicted feature vector with marks. 
These feature vectors are translation-invariant and stable to deformation of the domain.

Depending on the nature of the clustering of the points and the regularity of the marking function,
we find that the scattering moments predict as good as the baseline approach, showing that they capture well the 
geometric property. In case where marks depend on points in a non-local way, the scattering moment representation  seems  much better. 

Future directions to be explored include how to combine both approaches to
make better regression on the marks. Variations of the scattering moments~\cite{mallat2018phase}
 may help better capture the geometry of images. Replacing the linear regression
by others such as kernel regression or  neural network approaches are
also of potential interest. 
Also, the impact of regularity or clustering of the underlying point patterns on the quality of scattering representation has to better understood.

On application side,
 it is promising to combine geometry with local demand to predict the
 real loads to obtain an operational method for cell load prediction
for cellular networks.

\vspace{-1ex}
\addtocounter{section}{1}
\addcontentsline{toc}{section}{References}
%

\begin{thebibliography}{10}
\providecommand{\url}[1]{#1}
\csname url@samestyle\endcsname
\providecommand{\newblock}{\relax}
\providecommand{\bibinfo}[2]{#2}
\providecommand{\BIBentrySTDinterwordspacing}{\spaceskip=0pt\relax}
\providecommand{\BIBentryALTinterwordstretchfactor}{4}
\providecommand{\BIBentryALTinterwordspacing}{\spaceskip=\fontdimen2\font plus
\BIBentryALTinterwordstretchfactor\fontdimen3\font minus
  \fontdimen4\font\relax}
\providecommand{\BIBforeignlanguage}[2]{{%
\expandafter\ifx\csname l@#1\endcsname\relax
\typeout{** WARNING: IEEEtran.bst: No hyphenation pattern has been}%
\typeout{** loaded for the language `#1'. Using the pattern for}%
\typeout{** the default language instead.}%
\else
\language=\csname l@#1\endcsname
\fi
#2}}
\providecommand{\BIBdecl}{\relax}
\BIBdecl

\bibitem{gscatt}
S.~Mallat, ``Group invariant scattering,'' \emph{Commun. Pure Appl. Math.},
  vol. 65(10), pp. 1331--1398, 2012.

\bibitem{lbfgs}
R.~H. Byrd, P.~Lu, and J.~Nocedal, ``A limited memory algorithm for bound
  constrained optimization,'' \emph{SIAM J Sci. Stat. Comp.}, vol.~16, no.~5,
  pp. 1190--1208, 1995.

\bibitem{lbfgs2}
C.~Zhu, R.~H. Byrd, P.~Lu, and J.~Nocedal, ``Algorithm 778: {L-BFGS-B}: Fortran
  subroutines for large-scale bound-constrained optimization,'' \emph{ACM
  Trans. Math. Softw.}, vol.~23, no.~4, pp. 550--560, 1997.

\bibitem{blaszczyszyn2018stochastic}
B.~B{\l}aszczyszyn, M.~Haenggi, P.~Keeler, and S.~Mukherjee, \emph{Stochastic
  geometry analysis of cellular networks}.\hskip 1em plus 0.5em minus
  0.4em\relax Cambridge University Press, 2018.

\bibitem{siomina2012analysis}
I.~Siomina and D.~Yuan, ``Analysis of cell load coupling for {LTE} network
  planning and optimization,'' \emph{IEEE Trans. Wireless Comm.}, vol.~11,
  no.~6, pp. 2287--2297, 2012.

\bibitem{blaszczyszyn2014user}
B.~Blaszczyszyn, M.~Jovanovic, and M.~K. Karray, ``How user throughput depends
  on the traffic demand in large cellular networks,'' in \emph{IEEE
  WiOpt/SpaSWin}, 2014.

\bibitem{blaszczyszyn2016spatial}
B.~Blaszczyszyn, R.~Ibrahim, and M.~K. Karray, ``Spatial disparity of {QoS}
  metrics between base stations in wireless cellular networks.'' \emph{IEEE
  Trans. Comm.}, vol.~64, pp. 4381--4393, 2016.

\bibitem{blaszczyszyn:hal-01824986}
B.~B{\l}aszczyszyn and M.~K. Karray, ``{Performance analysis of cellular
  networks with opportunistic scheduling using queueing theory and stochastic
  geometry},'' hal-01824986, 2018.

\bibitem{awan2018robust}
D.~A. Awan, R.~L. Cavalcante, and S.~Stanczak, ``A robust machine learning
  method for cell-load approximation in wireless networks,'' in \emph{IEEE
  ICASSP}, 2018.

\bibitem{multi}
M.~Eickenberg, G.~Exarchakis, M.~Hirn, and S.~Mallat, ``Solid harmonic wavelet
  scattering: Predicting quantum molecular energy from invariant descriptors of
  3d electronic densities,'' in \emph{NIPS}, 2017.

\bibitem{srqe}
M.~Hirn, S.~Mallat, and N.~Poilvert, ``Wavelet scattering regression of quantum
  chemical energies,'' \emph{Multiscale Model. Simul.}, vol. 15(2), pp.
  827--863, 2017.

\bibitem{text}
S.~Mallat and L.~Sifre, ``Rotation, scaling and deformation invariant
  scattering for texture discrimination,'' in \emph{IEEE CVPR}, vol. 65(10),
  2013, pp. 1233--1240.

\bibitem{penrose2003weak}
M.~D. Penrose and J.~E. Yukich, ``Weak laws of large numbers in geometric
  probability,'' \emph{Ann. Appl. Probab.}, vol.~13, no.~1, pp. 277--303, 2003.

\bibitem{baryshnikov2005gaussian}
Y.~Baryshnikov and J.~E. Yukich, ``Gaussian limits for random measures in
  geometric probability,'' \emph{Ann. Appl. Probab.}, vol.~15, pp. 213--253,
  2005.

\bibitem{blaszczyszyn2016limit}
B.~B{\l}aszczyszyn, D.~Yogeshwaran, and J.~Yukich, ``Limit theory for geometric
  statistics of point processes having fast decay of correlations,''
  \emph{arXiv preprint arXiv:1606.03988}, 2016, to appear in {\em Ann. Probab.}

\bibitem{scatt}
J.~Bruna, S.~Mallat, E.~Bacry, and J.-F. Muzy, ``Intermittent process analysis
  with scattering moments,'' \emph{Ann. Stat.}, vol.~43, pp. 323--351, 2015.

\bibitem{dcx-clust}
B.~B{\l}aszczyszyn and D.~Yogeshwaran, ``On comparison of clustering properties
  of point processes,'' \emph{Adv. Appl. Probab.}, vol.~46, no.~1, pp. 1--21,
  2014.

\bibitem{blaszczyszyn2015clustering}
------, ``Clustering comparison of point processes, with applications to random
  geometric models,'' in \emph{Stochastic Geometry, Spatial Statistics and
  Random Fields}.\hskip 1em plus 0.5em minus 0.4em\relax Springer, 2015, pp.
  31--71.

\bibitem{daley2007introduction}
D.~J. Daley and D.~Vere-Jones, \emph{An introduction to the theory of point
  processes: volume II}.\hskip 1em plus 0.5em minus 0.4em\relax Springer, 2007.

\bibitem{antoine1993image}
J.-P. Antoine, P.~Carrette, R.~Murenzi, and B.~Piette, ``Image analysis with
  two-dimensional continuous wavelet transform,'' \emph{Signal processing},
  vol.~31, no.~3, pp. 241--272, 1993.

\bibitem{last2017lectures}
G.~Last and M.~Penrose, \emph{Lectures on the Poisson process}.\hskip 1em plus
  0.5em minus 0.4em\relax Cambridge University Press, 2017, vol.~7.

\bibitem{blaszczyszyn1995factorial}
B.~B{\l}aszczyszyn, ``Factorial moment expansion for stochastic systems,''
  \emph{Stoch. Proc. Appl.}, vol.~56, no.~2, pp. 321--335, 1995.

\bibitem{blaszczyszyn1997note}
B.~B{\l}aszczyszyn, E.~Merzbach, and V.~Schmidt, ``A note on expansion for
  functionals of spatial marked point processes,'' \emph{Stat. \& Probab.
  Letters}, vol.~36, no.~3, pp. 299--306, 1997.

\bibitem{robert2014machine}
C.~Robert, \emph{Machine learning, a probabilistic perspective}.\hskip 1em plus
  0.5em minus 0.4em\relax Taylor \& Francis, 2014.

\bibitem{baddeley2005spatstat}
A.~Baddeley, R.~Turner \emph{et~al.}, ``Spatstat: an {R} package for analyzing
  spatial point patterns,'' \emph{Journal of statistical software}, vol.~12,
  no.~6, pp. 1--42, 2005.

\bibitem{anden2014scatnet}
J.~And{\'e}n, L.~Sifre, S.~Mallat, M.~Kapoko, V.~Lostanlen, and E.~Oyallon,
  ``Scatnet,'' \url{http://www.di.ens.fr/data/software/scatnet}, 2014.

\bibitem{bruna2015super}
J.~Bruna, P.~Sprechmann, and Y.~LeCun, ``Super-resolution with deep
  convolutional sufficient statistics,'' \emph{arXiv:1511.05666}, 2015.

\bibitem{mallat2018phase}
S.~Mallat, S.~Zhang, and G.~Rochette, ``Phase harmonics and correlation
  invariants in convolutional neural networks,'' arXiv:1810.12136, 2018.

\end{thebibliography}
{\small 

}
%
%

\remove{

\appendix
}

\end{document}